\documentclass[sn-mathphys-num]{sn-jnl}% Math and Physical Sciences Numbered Reference Style 
\usepackage{graphicx}%
\usepackage{multirow}%
\usepackage{amsmath,amssymb,amsfonts}%
\usepackage{amsthm}%
\usepackage{mathrsfs}%
\usepackage[title]{appendix}%
\usepackage{xcolor}%
\usepackage{textcomp}%
\usepackage{manyfoot}%
\usepackage{booktabs}%
\usepackage{algpseudocode}%
\usepackage{listings}%

\usepackage{array}
\usepackage[ruled,linesnumbered]{algorithm2e}
\usepackage{float}

\theoremstyle{thmstyleone}%
\theoremstyle{thmstyletwo}%

\theoremstyle{thmstylethree}%

\begin{document}

\title[Article Title]{Tightly-Coupled VLP/INS Integrated Navigation by Inclination Estimation and Blockage Handling}


\author[1]{\fnm{Xiao} \sur{Sun}}\email{xsun@whu.edu.cn}

\author*[1,2]{\fnm{Yuan} \sur{Zhuang}}\email{yuan.zhuang@whu.edu.cn}

\author[1]{\fnm{Xiansheng} \sur{Yang}}\email{xiansheng.yang@whu.edu.cn}

\author[1]{\fnm{Jianzhu} \sur{Huai}}\email{jianzhu.huai@whu.edu.cn}

\author[1]{\fnm{Tianming} \sur{Huang}}\email{huangtianmingw@outlook.com}

\author[3]{\fnm{Daquan} \sur{Feng}}\email{fdquan@szu.edu.cn}

\affil*[1]{\orgdiv{State Key Laboratory of Information Engineering in Surveying, Mapping and Remote Sensing}, \orgname{Wuhan University}, \orgaddress{\city{Wuhan}, \country{China}}}

\affil[2]{\orgname{Wuhan Institute of Quantum Technology}, \orgaddress{\city{Wuhan}, \country{China}}}

\affil[3]{\orgdiv{Guangdong Province Engineering Laboratory for Digital Creative Technology}, \orgname{Shenzhen University}, \orgaddress{\city{Shenzhen}, \country{China}}}


\abstract{Visible Light Positioning (VLP) has emerged as a promising technology capable of delivering indoor localization with high accuracy. In VLP systems that use Photodiodes (PDs) as light receivers, the Received Signal Strength (RSS) is affected by the incidence angle of light, making the inclination of PDs a critical parameter in the positioning model. Currently, most studies assume the inclination to be constant, limiting the applications and positioning accuracy. Additionally, light blockages may severely interfere with the RSS measurements but the literature has not explored blockage detection in real-world experiments. To address these problems, we propose a tightly coupled VLP/INS (Inertial Navigation System) integrated navigation system that uses graph optimization to account for varying PD inclinations and VLP blockages. We also discussed the possibility of simultaneously estimating the robot's pose and the locations of some unknown LEDs. Simulations and two groups of real-world experiments demonstrate the efficiency of our approach, achieving an average positioning accuracy of 10 cm during movement and inclination accuracy within 1 degree despite inclination changes and blockages.}

\keywords{Visible light positioning (VLP), received signal strength (RSS), INS, inclination, blockage, graph optimization}



\maketitle

\section{Introduction}
\label{introduction}

Due to the increasing need for location-based service (LBS), Visible Light Positioning (VLP) has attracted great academic and commercial interests. VLP utilizes light-emitting diodes (LEDs) to emit light signals and shows great potential for indoor positioning systems (IPSs) since LEDs have many important features, such as high bandwidth, high energy efficiency, long lifetime, and low infrastructure cost \cite{8292854}. Compared with widely used IPS methods nowadays such as WiFi, Bluetooth, Ultra-Wideband (UWB), odometer, vision, and LiDAR, VLP shows multiple advantages \cite{8292854,el2021indoor} at the same time: high accuracy, low infrastructure and computational cost, and easy deployment, as listed in Table \ref{IPS}.

\begin{table*}[htbp]
	\centering
	\caption{Comparison of some IPSs.}  \label{IPS}
	\footnotesize
	\begin{tabular}{m{1cm}<{\centering}m{2cm}<{\centering}m{2cm}<{\centering}m{3cm}<{\centering}m{3cm}<{\centering}}
		\hline
		\textbf{Systems} & \textbf{Accuracy} &\textbf{Setup} & \textbf{Advantages} & \textbf{Drawbacks} \\
		\hline
		{\textbf{WiFi}}  &  {3-10 m} & {WiFi routers or access points} & {No additional infrastructure, low cost, wide coverage.} &  {Low accuracy.} \\
		\hline
		{\textbf{Bluetooth}} &  {1-5 m} & {BLE beacons} & {Low cost, low energy consumption, and easy to deploy.} &  {Low accuracy.} \\
		\hline
		{\textbf{UWB}} &  {Centimeters to decimeters} & {UWB tags} &  {High accuracy and insensitive to multipath effect.} &  {High infrastructure cost. } \\
		\hline
		{\textbf{Odometer}} &  {$\sim 1\%$ of path length} &  {Odometers} &  {Self-contained, anti-interference.} &  {Only determine forward and heading changes.}  \\
		\hline
		{\textbf{Vision}} &  {$\sim 1\%$ of path length} & {Cameras} &  {Not sensitive to multipath effect and no extra infrastructure needed.} &  {Susceptible to degradation and light.}\\
		\hline
		{\textbf{LiDAR}} &  {$\textless 1\%$ of path length} & {LiDAR sensors} &  {High accuracy, high time resolution, not sensitive to multipath effect.} &  {Large cost, susceptible to the weather.} \\
		\hline
		{\textbf{VLP}} &  {Centimeters to decimeters} & {LED lamps and light receivers} & {High accuracy, existing infrastructures, and low cost.} &  {Susceptible to signal blockage and light reflecting.} \\
		\hline
	\end{tabular}
\end{table*}

Among all the literature using LEDs for IPSs so far, Photodiodes (PDs) and cameras are the main receivers used in the VLP system. A PD converts the received light signals into current and outputs the luminance by measuring the voltage. In PD-based VLP systems, Received Signal Strength (RSS) \cite{9728724} is the most direct measurement that can be easily obtained by a single PD. By a combination of multiple PDs, Angle of Arrival (AOA) measurements are also available \cite{8675983, 9456850}. With additional devices, a PD can also measure Time of Arrival (TOA) \cite{7339418} and Time Difference of Arrival (TDOA) \cite{8368233} for positioning. Image-based VLP systems utilize imaging geometry to solve the cameras' position by taking photos of LEDs. Since cameras are commonly equipped sensors, image-based VLP is also used in both consumer products and industry \cite{9330552}. A majority of VLP systems \cite{8292854} use a PD as the receiver since the cost of a single PD is significantly lower. This paper focuses mainly on PD-based systems.

In general, PD-based positioning systems use the Lambertian radiation model to estimate the receiver's position. As Lambertian radiation is defined by the distance of light transmission, incident, and irradiation angles, positioning is difficult when the inclination of the PD varies. To simplify the positioning problem, most experimental \cite{9728724, 9547766, 2616479, 7997704} and part of simulated \cite{9184060, 016106} PD-based research assumed the receiver is levelly placed. The research \cite{6823667, 8675983, 8835022, 8106665} considered the tilt angle in the positioning model, but the tilt angles were known values. However, 3D positioning and inclination estimation are necessary for some users, e.g., pedestrians on the stairs, legged robots, and indoor UAVs.

To cope with the unknown inclinations of the receiver, plenty of works \cite{8322671, 6868970, Wang2018, 9523865} used Inertial Measurement Unit (IMU) sensors (including accelerometer and gyroscope) to estimate the inclination. Zou \textit{et al}. \cite{8322671} fused VLP and Inertial Navigation System (INS) using an Unscented Kalman Filter (UKF) to solve the tilt angle. However, their system wrongly assumed that the tilt axis was perpendicular to the Line of Sight (LOS) and reduced the geometric problem to a two-dimensional one. The works \cite{6868970, Wang2018} used an accelerometer while the work \cite{9523865} used the integration of a gyroscope's data to estimate the inclination of the receiver; additionally, Wang \textit{et al}. \cite{Wang2018} used a magnetometer to estimate the heading angle. However, the inclinations estimated from these methods may suffer from IMU's bias errors. Also, these positioning methods are not robust when facing Non-Line-Of-Sight (NLOS) situations (e.g. light blockages). Additionally, Zhou \textit{et al}. \cite{8486755, 8805277} used convex optimization to simultaneously estimate the receiver's location and orientation, but their simulation experiments show the demand for a sufficient number (at least 20) of LEDs in the Field Of View (FOV), which is not applicable in real applications.

Visible light cannot penetrate opaque objects, thus, LOS blockages can lead to wrong RSS measurements and severely damage the positioning system \cite{9728724,9721812,8891724,9184060}. Literature has discussed the theoretical calculation \cite{9226495, 9184905} and the solution \cite{9523865, 9721812, 8891724, 9184060, 9728724} of the blockages. Tang \textit{et al.} \cite{9226495} gave a calculation on shadow area caused by the blockage of a cylinder. Hosseinianfar and Brandt-Pearce \cite{9184905} proposed an optimization method to solve the location of a pedestrian who blocks the LOS signal by modeling the person as a cylinder. Sheikholeslami \textit{et al}. \cite{9523865} used a Convolutional Neural Network (CNN), along with inertial sensors, to predict the locations during blockages in simulation. Vuong \textit{et al}. \cite{9721812} combined trilateration-based RSS with NLOS-based fingerprinting model to deal with the potentially blocked situation. Yang \textit{et al}. \cite{8891724} used Pedestrian Dead Reckoning (PDR) to estimate the velocity and predicted locations during blockages. Younus \textit{et al}. \cite{9569377} discussed the impact of blocking on received power distribution and positioning but did not give a solution. Zhang \textit{et al}. \cite{9184060} proposed a partial-RSS-assisted inertial navigation system with a Recurrent Neural Network (RNN), but both their VLP and inertial model were simplified to a 2-Degree-of-Freedom (2DoF) situation. Our previous work \cite{9728724} used a robust graph optimization method to smooth the RSS data and resist the errors caused by short-time blockage but did not detect them. To our knowledge, there are no studies with real test data about blockage handling using RSS.

\begin{figure}[htpb]
	\centering
	\includegraphics[width=3in]{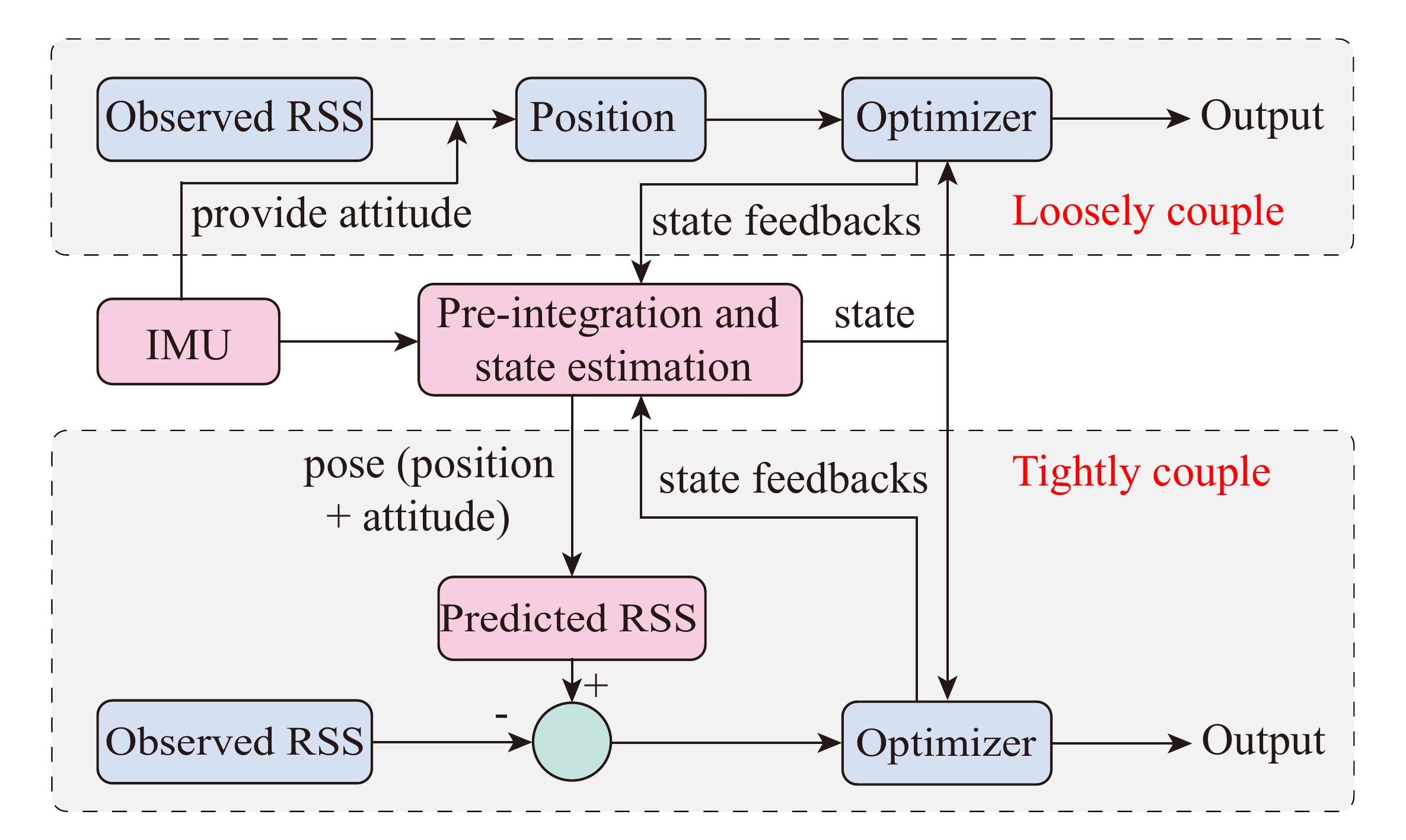}
	\caption{A flow chart showing the difference between tightly-coupled and loosely-coupled VLP/INS integration.}
	\label{tight_loose}
\end{figure}

To improve the robustness of the PD-based VLP system, it is a reasonable choice to be aided with INS. \textit{Tightly-coupled}, relative to the term \textit{loosely-coupled}, is a concept originated from the field of GNSS/INS integrated navigation \cite{RN20, li2021semi,ZHUANG202362}. Fig. \ref{tight_loose} shows differences between tightly \cite{9184060, 8322671} and loosely \cite{7997704} coupled VLP/INS systems used in previous studies. Both tightly and loosely coupled systems take advantage of the INS but tightly-coupled models can better resist IMU errors and can output continuous navigation results even when blockages happen. The research \cite{9184060} and \cite{8322671} are tightly designed. But the system in the work \cite{9184060} is 2-dimensional by assuming the receiver to be levelly placed; the model in \cite{8322671} wrongly assumed that the tilt axis was perpendicular to the LOS and reduces the geometric problem to a two-dimensional one. To the best of our knowledge, there is no 6DoF tightly-coupled VLP/INS system that solved the problems of both inclination estimation and blockage handling.

In this work, we propose a tightly coupled VLP/INS integrated navigation system that solves the problem of attitude (including inclination and heading) estimation and signal blockages. The contributions are as follows:
\begin{itemize}
	\item{We design a graph optimization model with a sliding window to optimize the position and inclination of a PD by fusing VLP and IMU data, where the pose (position and attitude) of a PD is used to calculate the Lambertian channel gain. }
	\item{We deduce the disturbance model of the Lambertian law to construct the Jacobian matrix during optimization and study the observability of the state using RSS measurements.}
	\item{To eliminate the LOS blockage, we propose a blockage detection method to pick out the interfered RSS measurements and exclude them.}
	\item{We discussed the feasibility of simultaneously estimating PD's pose and some unknown LEDs' locations.}
	\item{We implemented field vehicular tests using mobile robots assembled with low-cost PDs and IMUs in two scenarios. To our knowledge, we are the first to study LOS blockages in RSS-based VLP systems by real test data.}
\end{itemize}

We organize the rest of this article as follows. Section \ref{overview} introduces the basic structure and application scenario of our system and defines the coordinate systems. Section \ref{VLP} presents the quaternion-based Lambertian model, weight determination, and blockage detection, which are the main contributions of this article. The IMU pre-integration and graph optimization model are introduced in Section \ref{integration}. The experimental setup for VLP/INS integrated navigation is described in Section \ref{experiment}. The experimental results, analysis, and discussions are shown in Section \ref{result}, and the conclusions are presented in Section \ref{conclusion}.

\section{System Overview}
\label{overview}
\begin{figure}
	\centering
	\includegraphics[width=3.5in]{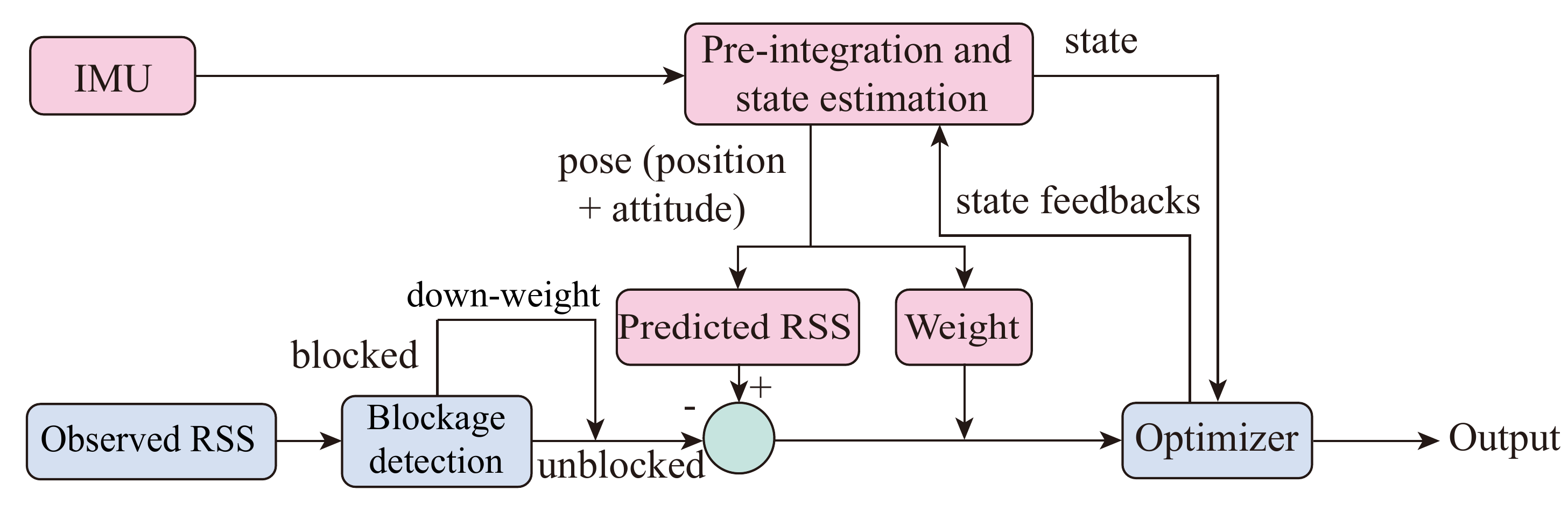}
	\caption{The overall architecture of the proposed tightly-coupled VLP/INS integrated navigation system.}
	\label{scheme}
\end{figure}

\begin{figure}
	\centering
	\includegraphics[width=3.5in]{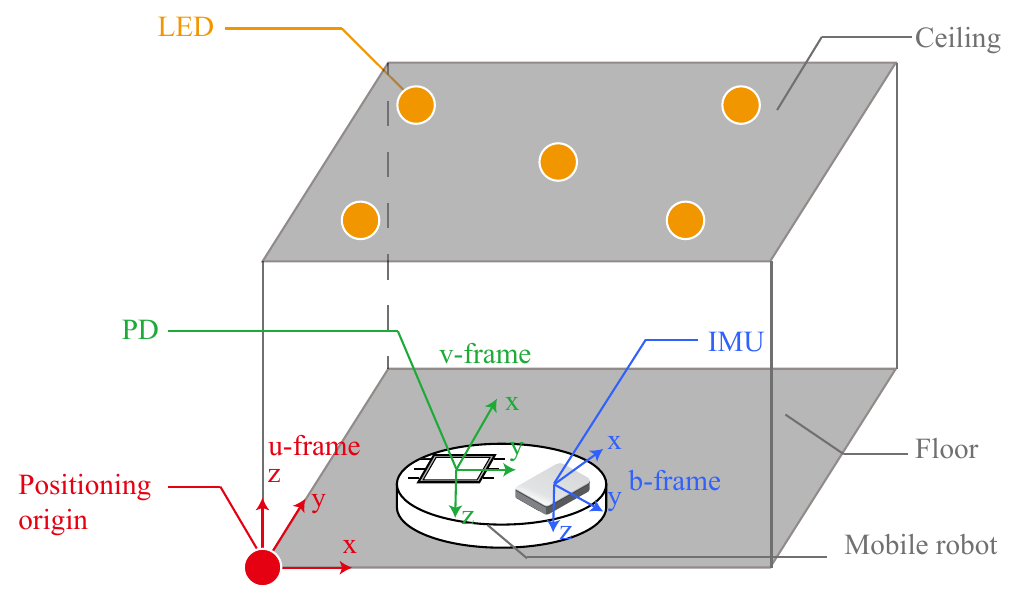}
	\caption{The coordinate systems defined in our system.}
	\label{coordinate}
\end{figure}

The structure of the proposed tightly-coupled VLP/INS integrated navigation system is shown in Fig. \ref{scheme}. Different from the traditional concept of tightly-coupled VLP/INS fusion system as Fig. \ref{tight_loose} shows, we add a blockage detection module to improve the accuracy and robustness. Fig. \ref{coordinate} shows the simplified application scene of our system, where LEDs are installed on the ceiling of a room and the receiver is moving on the floor. For this scene, we have the assumption that the height of a mobile robot is constant when it is operating on a fixed floor.

To start with, we present the notations used in this article. We define $(\cdot)^u$ as the indoor frame, whose origin is at the corner of a room, and axes are based on the orientation of the walls (drawn in red in Fig. \ref{coordinate}). $(\cdot)^b$ is the body frame, which we define based on the position and orientation of the IMU (drawn in blue in Fig. \ref{coordinate}). $b_k$ means the body frame at time $t_k$. $(\cdot)^v$ is the VLP frame, which we define based on the surface of the PD and the vehicle's travel direction (forward-right-down, drawn in green in Fig. \ref{coordinate}). $v_k$ means the VLP frame at time $t_k$.

\section{RSS-based Indoor Visible Light Positioning}
\label{VLP}
\subsection{Quaternion-based Lambertian Model}
\label{lambertian}
In this paper, we consider a system containing $N$ LEDs and a single-PD-based VLP receiver. A PD can collect optical signals simultaneously from several LEDs, forming a time series of amplitudes. To obtain RSS values, we distinguish optical signals from each LED based on its known modulated frequency. To this end, we adopt the Discrete Fourier Transform (DFT) to separate LED signals in the frequency domain and obtain the $l^{th}$ LED's RSS value $P_{l}$ by measuring the amplitude peak. Within the FOV of LEDs and the PD, the received optical power can be modeled with the Lambertian law \cite{554222}:
\begin{equation}
	\label{lamber}
	P_l=\frac{\left(m_l+1\right) A_R P_{Tl}}{2 \pi} \cdot \frac{\cos ^{m_l}(\theta) \cos(\psi) T_s(\psi) g(\psi)}{D_l^2}
\end{equation}
where $D_l$ is the distance between the $l^{th}$ LED and the PD, $A_R$ is the effective area of the PD, $P_{Tl}$ is the optical power of the $l^{th}$ LED, $\theta$ is the angle of irradiance from the $l^{th}$ LED, $\psi$ is the angle of incidence at the receiver. $T_s(\psi)$ is the gain of an optical filter, and $g(\psi)$ is the gain of an optical concentrator placed in front of the detector; in usual applications,  $T_s(\psi)=g(\psi)=1$. $m_l$ is the Lambertian order of the $l^{th}$ LED chip.

To calculate the cosine functions, we use the dot product of a unit vector and a LOS vector,
\begin{equation}
	\cos(\psi)=\frac{\mathbf{n}^u\cdot\mathbf{D}_l^u}{D_l}
\end{equation}
\begin{equation}
	\label{theta}
	\cos(\theta)=\frac{\mathbf{n}_{l}^u\cdot\mathbf{D}_l^u}{D_l}
\end{equation}
\begin{equation}
	D_l=||\mathbf{D}_l^u||
\end{equation}
where $\mathbf{n}^u$ is the normal of the receiver plane, $\mathbf{n}_{l}^u$ is the unit vector indicating the opposite radiating direction of the $l^{th}$ LED. Both $\mathbf{n}^u$ and $\mathbf{n}_{l}^u$ are taken the upward direction. For indoor positioning, LEDs are usually installed on top of the ceiling, $\mathbf{n}_{l}^u=[0,0,1]$. $\mathbf{D}_l^u$ is the three-dimensional vector indicating the direction from the PD to the $l^{th}$ LED. Thus, the equation (\ref{lamber}) can be refactored to be as,
\begin{equation}
	\label{lamber2}
	P_l=\frac{\left(m_l+1\right) A_R P_{Tl}}{2 \pi} \cdot \frac{ \left(\mathbf{n}^u\cdot\mathbf{D}_l^u\right) \left(\mathbf{n}_{l}^u\cdot\mathbf{D}_l^u\right)^{m_l}} {D^{3+m_l}_l}
\end{equation}
In this paper, we use rotation matrices $\mathbf{R}$ and Hamilton quaternion $\mathbf{q}$ to represent the attitude. $\mathbf{n}^u$ is determined based on the receiver's attitude,
\begin{equation}
	\label{npd}
	\mathbf{n}^u=\mathbf{R}^u_{v}\mathbf{n}^{v}
\end{equation}
where the rotation matrix $\mathbf{R}^u_{v}\in SO(3)$ is the Direction Cosine Matrix (DCM) converting from v-frame to u-frame, $\mathbf{n}^{v}=[0,0,1]$. To express $\mathbf{n}$ by a quaternion, for $\mathbf{q}^u_{v}=q_0+q_1\mathbf{i}+q_2\mathbf{j}+q_3\mathbf{k}$, we can transfer $\mathbf{q}^u_{v}$ to $\mathbf{R}^u_{v}$ based on the quaternion theory and have the following expression:
\begin{equation}
	\label{n_q}
	\mathbf{n}^u=\left[
	\begin{matrix}
		2\left(q_1 q_3+q_0 q_2\right)\\
		2\left(q_2 q_3-q_0 q_1\right)\\
		q_0^2-q_1^2-q_2^2+q_3^2
	\end{matrix}
	\right]
\end{equation}

\subsection{Disturbance Model and Observability}
\label{disturbance}
RSS measurements are relevant with both positions and attitudes, but not all state components are observable and worth estimation. To make it clear for this situation, we take full derivatives of equation (\ref{lamber2}) with position disturbance $d\mathbf{r}$ and angle disturbance $d\boldsymbol{\phi}$, 
\begin{equation}
	\label{derivative}
	\begin{aligned}
		dP_l=&P_l\frac{\mathbf{D}_l^u\times\mathbf{n}^u} {\mathbf{D}_l^u\cdot\mathbf{n}^u} \cdot d\boldsymbol{\phi} + P_l\left[-\frac{\mathbf{n}^u}{\mathbf{n}^u\cdot\mathbf{D}_l^u} \right.\\
		&-\frac{m_l\mathbf{n}_{l}^u}{\mathbf{n}_{l}^u\cdot\mathbf{D}_l^u}+\left.(3+m_l)\frac{\mathbf{D}_l^u}{D^{2}_l}\right] \cdot d\mathbf{r}
	\end{aligned}
\end{equation}
the deducing process is given in the Appendix \ref{secA1}. 

The equation (\ref{derivative}) illustrates how position and attitude are influenced by RSS errors, which can be caused by multi-path, NLOS, signal modulation, and noises \cite{8292854, yang2023deepwipos}. Indoor mobile robots usually move on a planar floor, therefore, we simplify (\ref{derivative}) to 2D situations. To make the optimizer concentrate more on planar positions, we fix the height of the PD during the optimization with RSS measurements, hence 3-dimensional $d\mathbf{r}$ is reduced to 2-dimensional $d\mathbf{s}$. Since $\mathbf{n}_{l}^u=[0,0,1]$, equation (\ref{derivative}) can be reduced to
\begin{equation}
	\label{derivative2}
	\begin{aligned}
		dP_l=&P_l\frac{\mathbf{D}_l^u\times\mathbf{n}^u} {\mathbf{D}_l^u\cdot\mathbf{n}^u} \cdot d\boldsymbol{\phi} + P_l\left[-\frac{\left(\mathbf{n}^u\right)_{xy}}{\mathbf{n}^u\cdot\mathbf{D}_l^u} \right.\\
		&+\left.(3+m_l)\frac{\mathbf{s}_l^u-\mathbf{s}^u}{D^{2}_l}\right] \cdot d\mathbf{s}
	\end{aligned}
\end{equation}
where $\left(\mathbf{n}^u\right)_{xy}$ is the $x$ and $y$ component of $\mathbf{n}^u$, $\mathbf{s}_l^u$ and $\mathbf{s}^u$ are the $x$ and $y$ components of the positions of the $l^{th}$ LED and the PD. In the equation (\ref{derivative2}), the height of a PD is unconstrained since $d\mathbf{s}$ is a planar position. To prevent height estimation from divergence, we design height constraints, which is demonstrated in Section \ref{optimization}.

Apart from height, the heading angle also needs to be dealt with specially. According to the Appendix, $d\boldsymbol{\phi}$ is the attitude disturbance expressed in the u-frame. Since $(\mathbf{D}_l^u\times\mathbf{n}^u) \cdot \mathbf{n}^u=0$ permanently, the heading angle is unobservable for a single VLP RSS measurement. There is an intuitive explanation for this phenomenon: when we rotate the plane of a PD, the RSS remains unchanged. Therefore, VLP-alone positioning can hardly solve the heading; other sensors' aiding (e.g. inertial sensors) is necessary. Additionally, for an IMU that does not provide magnetic measurements, the initial heading needs to be determined by external equipment.

\subsection{Blockage Detection}
\label{blockage}
Since visible light cannot penetrate opaque objects, light blockages can severely damage the RSS measurements. To prevent this from happening, blockage detection is necessary to exclude bad observations.
\begin{figure}[H]
	\centering
	\includegraphics[width=2in]{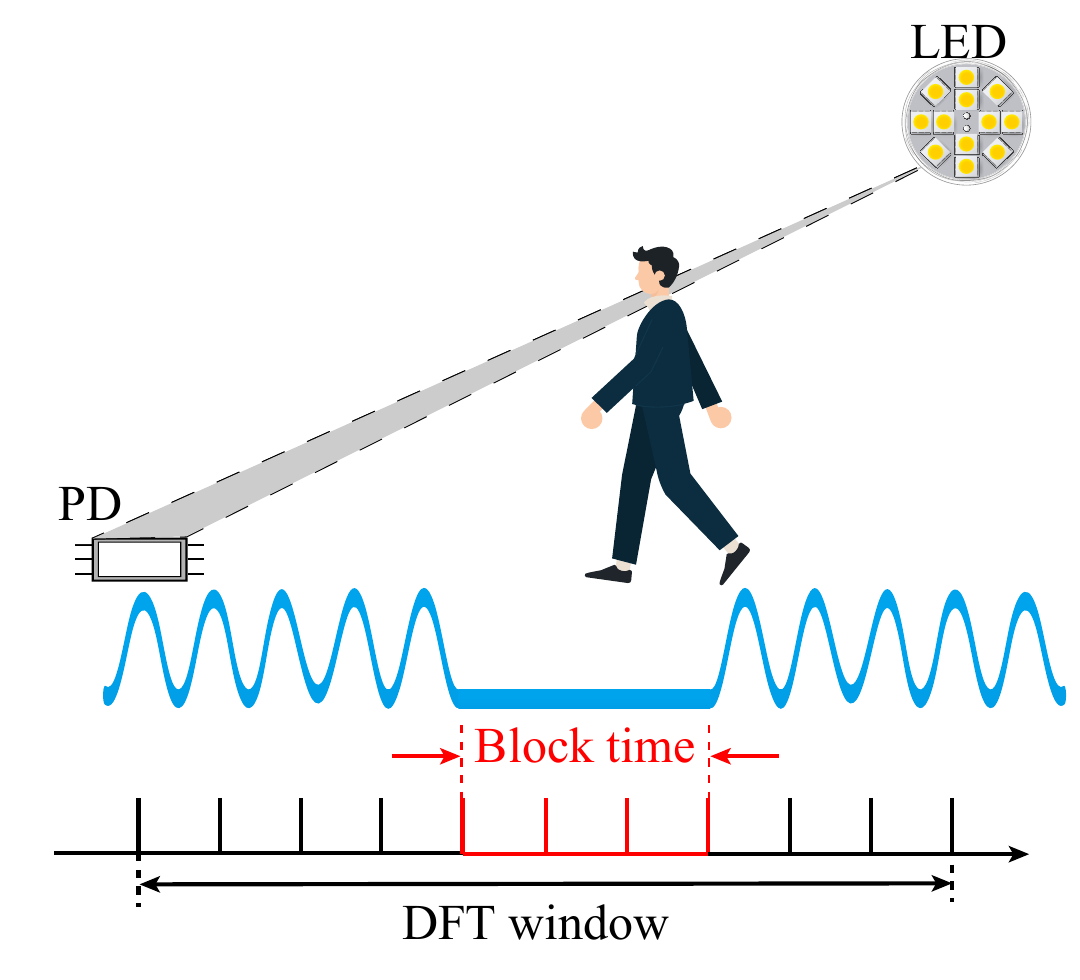}
	\caption{Illustration of the blockage for a sinusoidal-modulated wave and the influence on its DFT transformation.}
	\label{block}
\end{figure}

Suppose there is a reflection-free and interference-free environment, the RSS should be zero when blockages happen. However, in real applications, the RSS is calculated based on Discrete Fourier Transform (DFT) in a window of a certain data length, which can be explained by Fig. \ref{block}. Taking one-second voltage data, for example, if the blockages occupy $0.5 \mathrm{s}$ in this DFT window, the RSS measurement may descend to be half of the normal one, but not zero. Thus, we cannot simply detect blockages by finding zero measurements. In this paper, we proposed a Descending-Rising Detection (DRD) method to deal with raw PD observations.

\begin{figure}[H]
	\centering
	\includegraphics[width=3in]{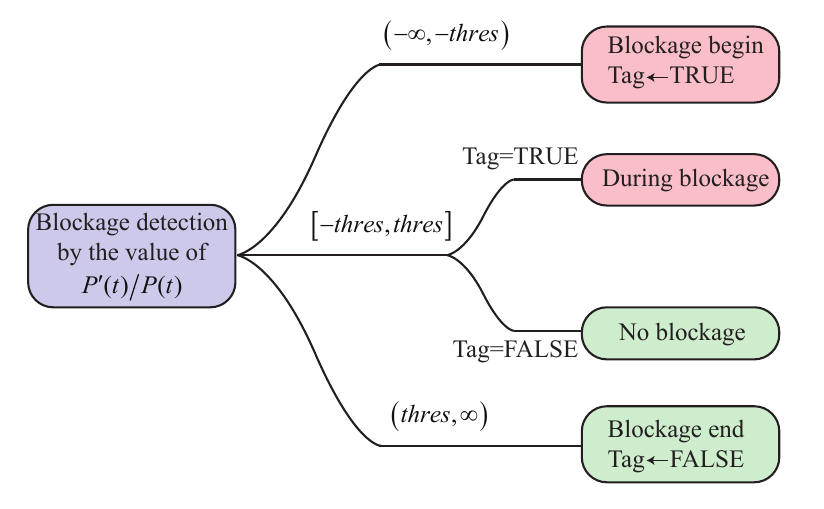}
	\caption{The blockage detection process by the value of RSS changing rate ratio.}
	\label{detect}
\end{figure}

Firstly, we introduce the basic principle: when blockages happen, RSS always suffers a severe descent; when blockages end, RSS rapidly rises to a normal level. The process is shown in Fig. \ref{detect}. In this algorithm, we use the RSS changing rate ratio threshold to separate blocked signal from unblocked one and use a blockage detection tag to indicate whether blockages are ongoing. To approximate the RSS changing rate ratio using discrete RSS data, it is more accurate to difference the high-frequency sampling ($\textgreater100\mathrm{Hz}$) RSS data,
\begin{equation}
	\label{RSS_rate}
	\frac{P^{'}(t_i)}{P(t_i)}=\frac{P(t_{i+1})-P(t_{i})}{\delta t_i P(t_i)}
\end{equation}
The critical part lies in the threshold to judge the RSS changing rate ratio. For 3-D positioning applications, based on equation (\ref{derivative}), we have
\begin{equation}
	\label{threshold1}
	\begin{aligned}
		\left|\left|\frac{P^{'}(t)}{P(t)}\right|\right|\le &\left|\left|\frac{\mathbf{D}_l^u\times\mathbf{n}^u} {\mathbf{D}_l^u\cdot\mathbf{n}^u} \right|\right| \omega_\mathrm{max} + \left|\left|-\frac{\mathbf{n}^u}{\mathbf{n}^u\cdot\mathbf{D}_l^u} \right.\right.\\
		&\left.\left.-\frac{{{m}_{l}}{{\mathbf{n}}_{l}^u}}{{{\mathbf{n}}_{l}^u}\cdot {{\mathbf{D}}_{l}^u}}+ (3+{{m}_{l}})\frac{{{\mathbf{D}}_{l}^u}}{D_{l}^{2}}\right|\right| v_\mathrm{max}
	\end{aligned}
\end{equation}
where $\omega_\mathrm{max}$ and $v_\mathrm{max}$ are the maximum possible angular rate and velocity determined by vehicle state at that moment. The right side of the inequality (\ref{threshold1}) can be used for the threshold. In a 2-D situation when there is no attitude variations, inequality (\ref{threshold1}) can be reduced to
\begin{equation}
	\label{threshold2}
	\left|\left|\frac{P^{'}(t)}{P(t)}\right|\right|\le \frac{(3+m_l)s}{s^2+h^2}v_\mathrm{max}
\end{equation}
where $s$ and $h$ are the horizontal and vertical distances between an LED and a PD. To accurately get the time derivatives of RSS $P^{'}(t)$, we need to differentiate the RSS measurements with a high sampling rate, e.g., in our experiments, 120Hz.

Blocked RSS measurements can be either excluded or significantly down-weighted when constructing the optimization cost function. The down-weighting strategy is demonstrated in Section \ref{optimization}.

\section{Tightly-coupled VLP/INS Positioning}
\label{integration}
\subsection{IMU Pre-Integration and Residuals}
To connect neighboring states, we integrate IMU measurements during a certain time interval using the pre-integration method. Since the sampling rate of VLP measurements is always much lower than that of IMU, IMU pre-integration is a useful tool to make the IMU residuals and VLP residuals aligned in time. The commonly used IMU pre-integration model \cite{RN297} integrates accelerometer and gyroscope measurements within the time interval $\left[t_k,t_{k+1}\right]$ in the body frame. In our VLP/INS integration scheme, we compute all states in the VLP frame. To realize this, we use the DCM $\mathbf{R}_b^v$ to transform the $i^{th}$ acceleration $\hat{\mathbf{a}}_i^b$, angular rate $\hat{\boldsymbol{\omega}}_i^b$ measurements and their biases from b-frame to v-frame, and then calculate the pre-integration vector $\hat{\mathbf{z}}_{v_{k+1}}^{v_{k}}=\left[\hat{\boldsymbol{\alpha}}_{v_{k+1}}^{v_{k}},\hat{\boldsymbol{\beta}}_{v_{k+1}}^{v_{k}},\hat{\boldsymbol{\gamma}}_{v_{k+1}}^{v_{k}}\right]$. The DCM $\mathbf{R}_b^v$ can be determined by the installation angles of the IMU and PD.

\begin{figure}[H]
	\centering
	\includegraphics[width=3in]{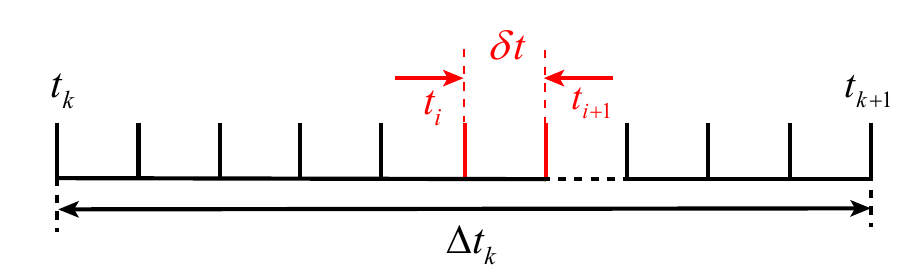}
	\caption{IMU sample time and pre-integration interval.}
	\label{preintegration_fig}
\end{figure}

The relationship among time stamps is drawn in Fig. \ref{preintegration_fig}. In each time interval $[t_i,t_{i+1}]$, there are one accelerometer and one gyroscope measurements. Approximately, we assume the reference frame on each time $t_i$ to be the VLP frame $v_k$. The equations (\ref{preintegration}) present the propagation formula to calculate step by step \cite{RN297}.
\begin{equation}
	\label{preintegration}
	\begin{aligned}
		\hat{\boldsymbol{\alpha}}_{i+1}^{v_{k}}&=\hat{\boldsymbol{\alpha}}_{i}^{v_{k}}+\hat{\boldsymbol{\beta}}_{i}^{v_{k}}\delta t+\frac{1}{2}\mathbf{R}(\hat{\boldsymbol{\gamma}}^{v_k}_i)(\hat{\mathbf{a}}_i-\mathbf{b}_{a_i})\delta t^2\\
		\hat{\boldsymbol{\beta}}_{i+1}^{v_{k}}&=\hat{\boldsymbol{\beta}}_{i}^{v_{k}}+\mathbf{R}(\hat{\boldsymbol{\gamma}}^{v_k}_i)(\hat{\mathbf{a}}_i-\mathbf{b}_{a_i})\delta t\\
		\hat{\boldsymbol{\gamma}}_{i+1}^{v_{k}}&=\hat{\boldsymbol{\gamma}}_{i}^{v_{k}}\otimes\left[\begin{matrix}1\\\frac{1}{2}(\hat{\boldsymbol{\omega}}_i-\mathbf{b}_{g_i})\delta t\end{matrix}\right]
	\end{aligned}
\end{equation}
where $\delta t$ is the time difference between $t_i$ and $t_{i+1}$, $\mathbf{b}_{a_i}$ and $\mathbf{b}_{g_i}$ are IMU biases, $\mathbf{R}(\cdot)$ is calculated based on quaternions.

In our tightly-coupled RSS-based VLP/INS integrated navigation system, we choose a 16-dimensional vector to form the state,
\begin{equation}
	\mathbf{x}=\left[\begin{matrix}(\mathbf{p}^u)^T & (\mathbf{v}^u)^T & (\mathbf{q}^u_{v})^T & (\mathbf{b}_{a})^T & (\mathbf{b}_{g})^T\end{matrix}\right]^T
\end{equation}
Using the states at time $t_k$ and $t_{k+1}$, the residual for pre-integrated IMU measurement can be defined based on the pre-integration $\hat{\mathbf{z}}_{b_{k+1}}^{b_{k}}$ and states $ \boldsymbol{\mathcal{X}}=\left[ \mathbf{x}_0,\mathbf{x}_1,\ldots,\mathbf{x}_{n} \right] $ within the sliding window \cite{RN297}:
\begin{equation}
	\label{rb}
	\begin{aligned}
		&\mathbf{r}_{\mathcal{B}}\left(\hat{\mathbf{z}}_{v_{k+1}}^{v_{k}},\boldsymbol{\mathcal{X}}\right)=\left[
		\begin{matrix}
			\delta \boldsymbol{\alpha}_{v_{k+1}}^{v_{k}}\\
			\delta \boldsymbol{\beta}_{v_{k+1}}^{v_{k}}\\
			\delta \boldsymbol{\theta}_{v_{k+1}}^{v_{k}}\\
			\delta \mathbf{b}_{a}\\
			\delta \mathbf{b}_{g}\\
		\end{matrix}\right]\\
		&=\left[\begin{matrix}
			\mathbf{R}^{v_k}_u(\mathbf{p}^u_{k+1}-\mathbf{p}^u_{k}+\frac{1}{2}\mathbf{g}^u\Delta t^2_k-\mathbf{v}^u_{k}\Delta t_k)-\hat{\boldsymbol{\alpha}}_{v_{k+1}}^{v_{k}}\\
			\mathbf{R}^{v_k}_u(\mathbf{v}^u_{k+1}+\mathbf{g}^u\Delta t_k-\mathbf{v}^u_{k})-\hat{\boldsymbol{\beta}}_{v_{k+1}}^{v_{k}}\\
			2\left[({\mathbf{q}^u_{v_k}})^{-1}\otimes\mathbf{q}^u_{v_{k+1}}\otimes\left(\hat{\boldsymbol{\gamma}}_{v_{k+1}}^{v_{k}}\right)^{-1}\right]_{xyz}\\
			\mathbf{b}_{av_{k+1}}-\mathbf{b}_{av_{k}}\\
			\mathbf{b}_{gv_{k+1}}-\mathbf{b}_{gv_{k}}
		\end{matrix}\right]
	\end{aligned}
\end{equation}
where $\mathbf{R}^{v_k}_u$ is the DCM matrix to transform vectors from u-frame to v-frame, $\mathbf{g}^u$ is the gravity vector in u-frame, $\Delta t_k$ is the time difference of the time interval $\left[t_k,t_{k+1}\right]$, $[\cdot]_{xyz}$ means the vector constructed by the imaginary part of a quaternion. The covariance of $\mathbf{r}_{\mathcal{B}}\left(\cdot\right)$ is propagated by:
\begin{equation}
	\label{cov_v}
	\boldsymbol{\Sigma}_{i+1}^{v_k}=\boldsymbol{\Phi}_{i+1}\boldsymbol{\Sigma}_{i}^{v_k}\boldsymbol{\Phi}_{i+1}^T+\boldsymbol{Q}_{i+1},
\end{equation}
until the time $t_{i+1}$ is equal to $t_{k+1}$. The calculation of matrices $\boldsymbol{\Phi}_{i+1}$ and $\boldsymbol{Q}_{i+1}$ can be referred to \cite{9841461}.

\subsection{Graph Optimization for VLP/INS Integrated Navigation}
\label{optimization}
\begin{figure}[htpb]
	\centering
	\includegraphics[width=3.5in]{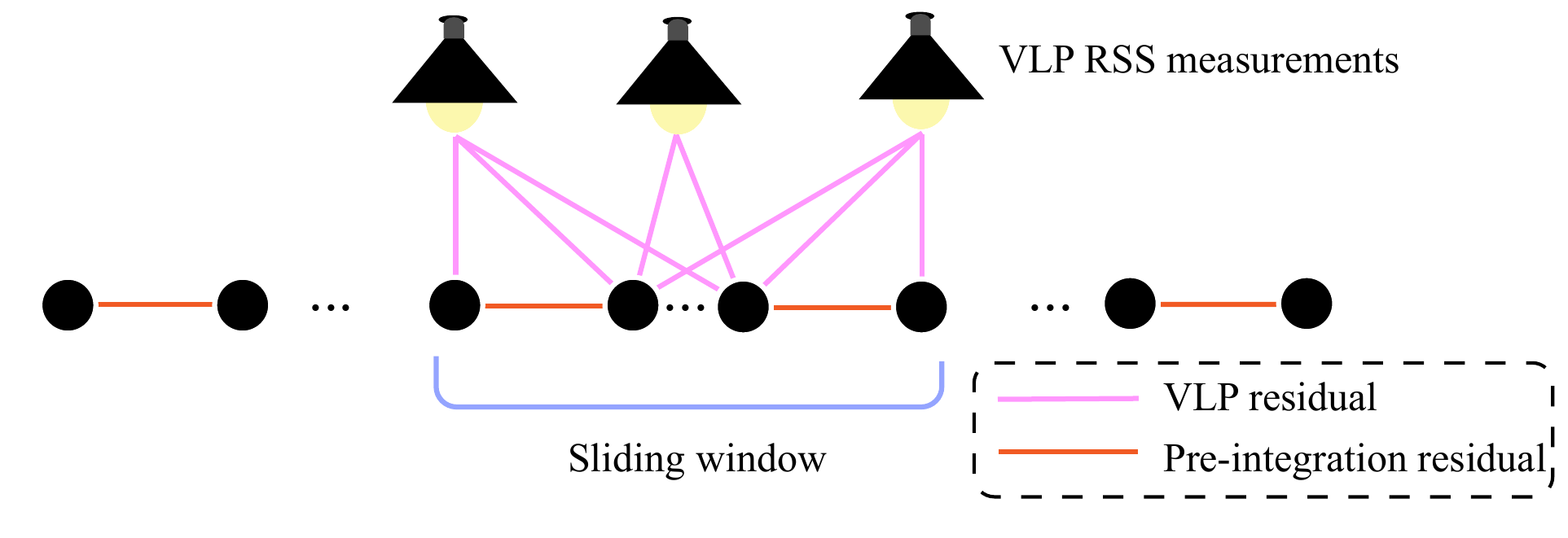}
	\caption{The graph structure of tightly-coupled VLP/INS fusion.}
	\label{graphopti}
\end{figure}

As introduced in Section \ref{introduction}, our tightly-coupled VLP/INS navigation method fused IMU measurements with VLP RSS observations, rather than VLP positioning results. To find a globally optimal navigation solution, we used a sliding window to optimize IMU and RSS measurements within a time interval of a suitable length, which is demonstrated in Fig. \ref{graphopti}. In real-time applications, the sliding window can slide to include the latest measurements and output the latest navigation results. The overall algorithm is demonstrated in Algorithm \ref{alg:alg3}. Similar to the pre-integration factor, we calculate the VLP residual based on measurements $\hat{P}_l^k$ and states $\boldsymbol{\mathcal{X}}$:
\begin{equation}
	\label{rv}
	\begin{aligned}
		\mathbf{r}_{\mathcal{V}}\left(\hat{P}_l^k, \boldsymbol{\mathcal{X}}\right) =&\frac{\left(m_l+1\right) A_R P_{Tl}}{2 \pi} \cdot \frac{ \left(\mathbf{n}^u\cdot\mathbf{D}_l^u\right) \left(\mathbf{n}_{l}^u\cdot\mathbf{D}_l^u\right)^{m_l}} {D^{3+m_l}_l}\\
		& -\hat{P}_l^k
	\end{aligned}
\end{equation}
where $\mathbf{D}_l^u$, $\mathbf{n}^u$, and $D_l$ are determined by the states $\boldsymbol{\mathcal{X}}$, which is stated in Section \ref{lambertian}. When computing $\mathbf{D}_l^u$, the state $\mathbf{p}^u$ needs to be corrected by the lever-arm since the PD and the IMU are not at the same place:
\begin{equation}
	\label{sec4:leverarm}
	\hat{\mathbf{p}}^u = \mathbf{p}^u + \mathbf{C}^{u}_v\mathbf{C}^{v}_b \boldsymbol\ell^{b}
\end{equation}
where $\boldsymbol\ell^{b}$ is the lever-arm from the IMU center to the PD center in the b-frame. Under the condition when the relative pose between the PD and the IMU is fixed, $\mathbf{C}^{v}_b$ is constant. But the corresponding quaternion of $\mathbf{C}^{u}_v$ has been defined as a part of the state, therefore, the lever-arm contribution to its disturbance needs to be taken into account. We have \cite{chen2021semi}:
\begin{equation}
	\label{leverarm_corr}
	\hat{\mathbf{C}}^{u}_v\mathbf{C}^{v}_b \boldsymbol\ell^{b}=\mathbf{C}^{u}_v\mathbf{C}^{v}_b \boldsymbol\ell^{b}+\left[\left(\mathbf{C}^{u}_v\mathbf{C}^{v}_b \boldsymbol\ell^{b}\right)\times\right]\delta\boldsymbol{\phi}
\end{equation}
The definition of antisymmetric matrix is referred to as \cite{sola2017quaternion}.

The disturbance model (\ref{derivative2}) provides partial derivatives of RSS on position and attitude, where the attitude is expressed in the form of Euler angle. To suit the state expressed with quaternions, we have the differentiation \cite{9841461, sola2017quaternion}:
\begin{equation}
	\label{diff_q}
	d\mathbf{q}_v^u=\frac{1}{2}\mathbf{q}_v^u\otimes\left[
	\begin{matrix}
		0\\
		d\boldsymbol{\phi}^v_{wv}\\
	\end{matrix}\right]
\end{equation}
In the state optimization process, it is a common practice to optimize a quaternion's minimal space, i.e., a 3-dimensional angle vector \cite{9841461, RN297}, and then update quaternions with manifold-form multiplier (\ref{diff_q}). Based on the equation (\ref{derivative2}), the Jacobian component of an RSS to a quaternion can be expressed as:
\begin{equation}
	\label{P_q}
	\begin{aligned}
		\frac{\partial P_l}{\partial \mathbf{q}_v^u}&=\frac{\partial P_l}{\partial \boldsymbol{\phi}^v_{wv}}=\frac{\partial P_l}{\partial \boldsymbol{\phi}^u_{vw}}\frac{\partial \boldsymbol{\phi}^u_{vw}}{\partial \boldsymbol{\phi}^v_{wv}}\\
		&=-P_l\frac{(\mathbf{D}_l^u\times\mathbf{n}^u)^T} {\mathbf{D}_l^u\cdot\mathbf{n}^u} \mathbf{R}_v^u
	\end{aligned}
\end{equation}

Based on (\ref{derivative}), (\ref{leverarm_corr}) and (\ref{P_q}), the Jacobian matrix for one VLP residual is calculated as follows:
\begin{equation}
	\label{jacobi}
	\begin{aligned}
		\mathbf{H}_{1\times16}&=\left[\hat{P}_l^k\left[-\frac{\mathbf{n}^u}{\mathbf{n}^u\cdot\mathbf{D}_l^u} -\frac{m_l\mathbf{n}_{l}^u}{\mathbf{n}_{l}^u\cdot\mathbf{D}_l^u}+(3+m_l)\frac{\mathbf{D}_l^u}{D^{2}_l}\right]^T, \mathbf{0}_{1\times4},-\hat{P}_l^k\frac{(\mathbf{D}_l^u\times\mathbf{n}^u)^T} {\mathbf{D}_l^u\cdot\mathbf{n}^u} \mathbf{C}_{v_k}^u\right.\\
		&\left. + \hat{P}_l^k\left[-\frac{\mathbf{n}^u}{\mathbf{n}^u\cdot\mathbf{D}_l^u} -\frac{m_l\mathbf{n}_{l}^u}{\mathbf{n}_{l}^u\cdot\mathbf{D}_l^u}+(3+m_l)\frac{\mathbf{D}_l^u}{D^{2}_l}\right]^T\left[\left(\mathbf{C}^{u}_v\mathbf{C}^{v}_b \boldsymbol\ell^{b}\right)\times\right],
		\mathbf{0}_{1\times6}\right]
	\end{aligned}
\end{equation}
As for 2-D situations, the Jacobian matrix can be calculated based on (\ref{derivative2}) and (\ref{P_q}). In real applications, $\mathbf{H}$ is a $N\times16$ matrix, where $N$ is the number of LEDs at each time stamp.

\begin{algorithm}[!h]
	\SetAlgoLined
	\KwIn{the state vector in a sliding window with width $n+1$: $ \boldsymbol{\mathcal{X}}=\left[ \mathbf{x}_0,\mathbf{x}_1,\ldots,\mathbf{x}_{n} \right] $; IMU pre-integration between the frames at the time $ t_k $ and $ t_{k+1} $: $ \hat{\mathbf{z}}_{v_{k+1}}^{v_k} $; RSS measurement of LED $l$ at time $t_k$: $ \hat{P}_l^k $}
	\KwOut{the optimized states: $ \hat{\boldsymbol{\mathcal{X}}}$}
	\BlankLine
	initialize position, velocity, attitude (PVA), and IMU errors\;
	\While{there are measurements}{
		integrate the IMU measurements using equation (\ref{rb}) within the time interval $[t_{k-1},t_k]$ to form $\hat{\mathbf{z}}_{v_{k+1}}^{v_k}$\;
		construct the cost function (\ref{cost}) within a sliding window whose width is $n+1$\;
		optimize using the LM method to obtain the optimal state vector $\boldsymbol{\mathcal{X}}$\;
		slide the window and marginalize the old variables;
	}
	\caption{The tightly-coupled VLP/INS graph optimization algorithm.}\label{alg:alg3}
\end{algorithm}
%\begin{algorithm}[H]
%	\caption{The tightly-coupled VLP/INS graph optimization algorithm.}\label{alg:alg3}
%	\begin{algorithmic}
%		\STATE \textbf{Parameter Statements:}
%		\STATE \hspace{0.25cm}$ \boldsymbol{\mathcal{X}}=\left[ \mathbf{x}_0,\mathbf{x}_1,\ldots,\mathbf{x}_{n} \right] $: the full state vector in a sliding window with width $n+1$.
%		\STATE \hspace{0.25cm}$ \hat{\mathbf{z}}_{v_{k+1}}^{v_k} $: IMU pre-integration between the frames at the time $ t_k $ and $ t_{k+1} $.
%		\STATE \hspace{0.25cm}$ \hat{P}_l^k $: RSS measurement of LED $l$ at time $t_k$.
%		\STATE \textbf{Initialization:}
%		\STATE \hspace{0.25cm}$ \mathbf{x}_{0} $: initial position, velocity, attitude (PVA), and IMU errors.
%		\STATE \textbf{Repeat:}
%		\STATE 1. IMU pre-integration: integrate the IMU measurements using equation (\ref{preintegration}) within the time interval $[t_{k-1},t_k]$ to form $\hat{\mathbf{z}}_{v_{k+1}}^{v_k}$.
%		\STATE 2. Construct the cost function (\ref{cost}) within a sliding window whose width is $n+1$. The weights of residuals are referred to in equations (\ref{cov_v}). The Jacobian matrix of a VLP residual is expressed in (\ref{jacobi}).
%		\STATE 3. Optimize using the Levenberg-Marquardt method \cite{RN380} to obtain the optimal state vector $\boldsymbol{\mathcal{X}}$.
%		\STATE 4. Compute PVA based on the state vector $\boldsymbol{\mathcal{X}}$ when it is necessary to output.
%		\STATE 5. Slide the window until there are no new measurements.
%		\STATE \textbf{End}
%	\end{algorithmic}
%\end{algorithm}

We minimize the sum of the Mahalanobis norm of all measurement residuals to obtain a maximum posterior estimation:
\begin{equation}
	\label{cost}
	\begin{aligned}
		\min \limits_{\boldsymbol{\mathcal{X}}}&\left\{\sum\limits_{k=0}^{n-1}\left\|\mathbf{r}_{\mathcal{B}}\left(\hat{\mathbf{z}}_{v_{k+1}}^{v_k}, \boldsymbol{\mathcal{X}}\right)\right\|_{\boldsymbol{\Sigma}_{v_{k+1}}^{v_k}}^2+\sum\limits_{k=0}^{n} \sum\limits_{l=0}^{m-1}\left\|\mathbf{r}_{\mathcal{V}}\left(\hat{P}_l^k, \boldsymbol{\mathcal{X}}\right)\right\|_{\Sigma_l}^2\right.\\
		&+\left.\sum\limits_{k=0}^{n} \left\|\mathbf{r}_{\mathcal{C}}\left(\mathbf{z}_{k}^u, \boldsymbol{\mathcal{X}}\right)\right\|_{\Sigma_k}^2\right\}
	\end{aligned}
\end{equation}
where $\mathbf{r}_{\mathcal{B}}\left(\cdot\right)$ and  $\mathbf{r}_{\mathcal{V}}\left(\cdot\right)$ are calculated by equations (\ref{rb}) and (\ref{rv}), $\Sigma_l$ is the covariance of $l^\mathrm{th}$ LED's RSS, for blocked measurements and signals out of FOV, its covariance can be set as a large value; in our test, we empirically set it as 99. $\mathbf{r}_{\mathcal{C}}\left(\cdot\right)$ is the constraint that can be non-holonomic constraints (NHC) \cite{7247673} or height constraints. In the 2-D positioning model, height is fixed and not observable in the VLP system (mentioned in Section \ref{disturbance}), thus, it needs to be constrained as:
\begin{equation}
	\label{constraint}
	\mathbf{r}_{\mathcal{C}}\left(\mathbf{z}_{k}^u, \boldsymbol{\mathcal{X}}\right)=(\mathbf{p}^u)_z-h_{\mathrm{PD}}
\end{equation}
where $h_{\mathrm{PD}}$ is the height of the PD that should be measured in advance. NHC is effective for a robot without lateral velocity while height constraint is effective in 2-D positioning situations. In our simulation (3-D), NHC is used, while in physical experiments (2-D), both NHC and height constraints are used.

To obtain the optimal state vector $\boldsymbol{\mathcal{X}}$, we use the Levenberg-Marquardt algorithm \cite{RN380}. In real-time applications, we should output the last state $\mathbf{x}_{n}$ within the sliding window.

\subsection{Unknown LED parameters}
\label{sec_unknown}
Apart from pose graph optimization, it is feasible to simultaneously estimate the poses and the parameters of some unknown LEDs in our graph optimization system. Since these parameters remain constant throughout the whole measurements, setting these parameters to be unknown will not add too much burden to the whole optimization problem. When some LEDs' locations are unknown, the state $\boldsymbol{\mathcal{X}}$ need to be extended to $\left[\boldsymbol{\mathcal{X}}, \boldsymbol{\mathcal{P}}\right]$, where $\boldsymbol{\mathcal{P}}$ is the set of LED locations that needs estimation; thereby the VLP residual function should be changed to $\mathbf{r}_{\mathcal{V}}\left(\hat{P}_l^k, \boldsymbol{\mathcal{X}}, \boldsymbol{\mathcal{P}}\right)$. To optimize the LED locations, we calculated the Jacobian matrix of $\mathbf{r}_{\mathcal{V}}$ over $\boldsymbol{\mathcal{P}}$ based on equation (\ref{rv}):
\begin{equation}
	\mathbf{H}_{\mathrm{LED},1\times2}=\hat{P}_l^k\left[\frac{\left(\mathbf{n}^u\right)_{xy}}{\mathbf{n}^u\cdot\mathbf{D}_l^u}-(3+m_l)\frac{\mathbf{s}_l^u-\mathbf{s}^u}{D^{2}_l}\right]^T
\end{equation}

The precision of LED locations depends partly on the motion of a robot. In a situation when the location of one LED is unknown, the discrete positions of a moving PD can be regarded as signal transmitters and the unknown LED can be regarded as a receiver. On the condition the PD-LED direction changes dramatically, the LED positions can be accurately estimated; otherwise, it is difficult to solve them. This situation can be quantified by a concept, Dilution of Precision (DOP) \cite{santerre_geometry_2017}, which originated from the Loran-C navigation system and the GNSS. In our graph optimization scheme, a position sequence within a sliding window can be used to locate the unknown LEDs. In this model, the precision of the LED position is determined by the trajectory's precision and the DOP value.

\begin{figure}[H]
	\centering
	\includegraphics[width=3.5in]{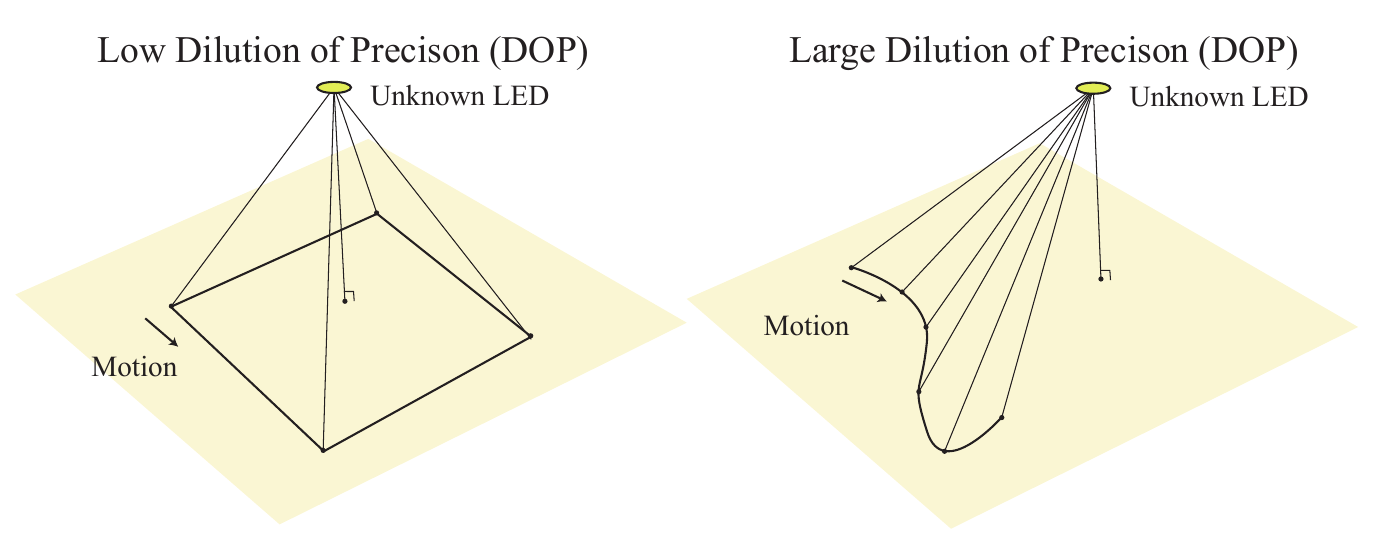}
	\caption{When a user trajectory is used to locate an unknown LED, the localization precision is determined by the trajectory's precision and the DOP value. Scattered trajectory points have strong geometry and a low DOP value, while aggregated points have a large DOP value. If two trajectories have the same precision, one with a low DOP value will lead to high localization precision of the unknown LED.}
	\label{DOP}
\end{figure}

The idea of geometric DOP is to state how errors in the measurement will affect the final state estimation. In a trilateration problem, when the known points are far apart, the geometry is strong and the DOP value is low (shown in Fig. \ref{DOP}). The mathematical expression of DOP can be referred to in \cite{santerre_geometry_2017}. In Section \ref{unknown}, we will present some test results where the locations of some LEDs are unknown.

\section{Experiment Setup}
\label{experiment}
We implemented both simulation and real-world experiments to test our tightly-coupled VLP/INS graph optimization system: the simulation is to exclude indoor errors and evaluate 3D positioning performance while real experiments are to estimate 2D positions and inclinations. The real-world experiments are implemented in two sites, both are empty rooms, one with the size of $3.8 \mathrm{m} \times 6.3 \mathrm{m} \times 2.8 \mathrm{m}$ and the other is $6 \mathrm{m} \times 8.4 \mathrm{m} \times 2.8 \mathrm{m}$, which are marked as experiment A and B, respectively. 

\begin{figure}[H]
	\centering
	\includegraphics[width=3.5in]{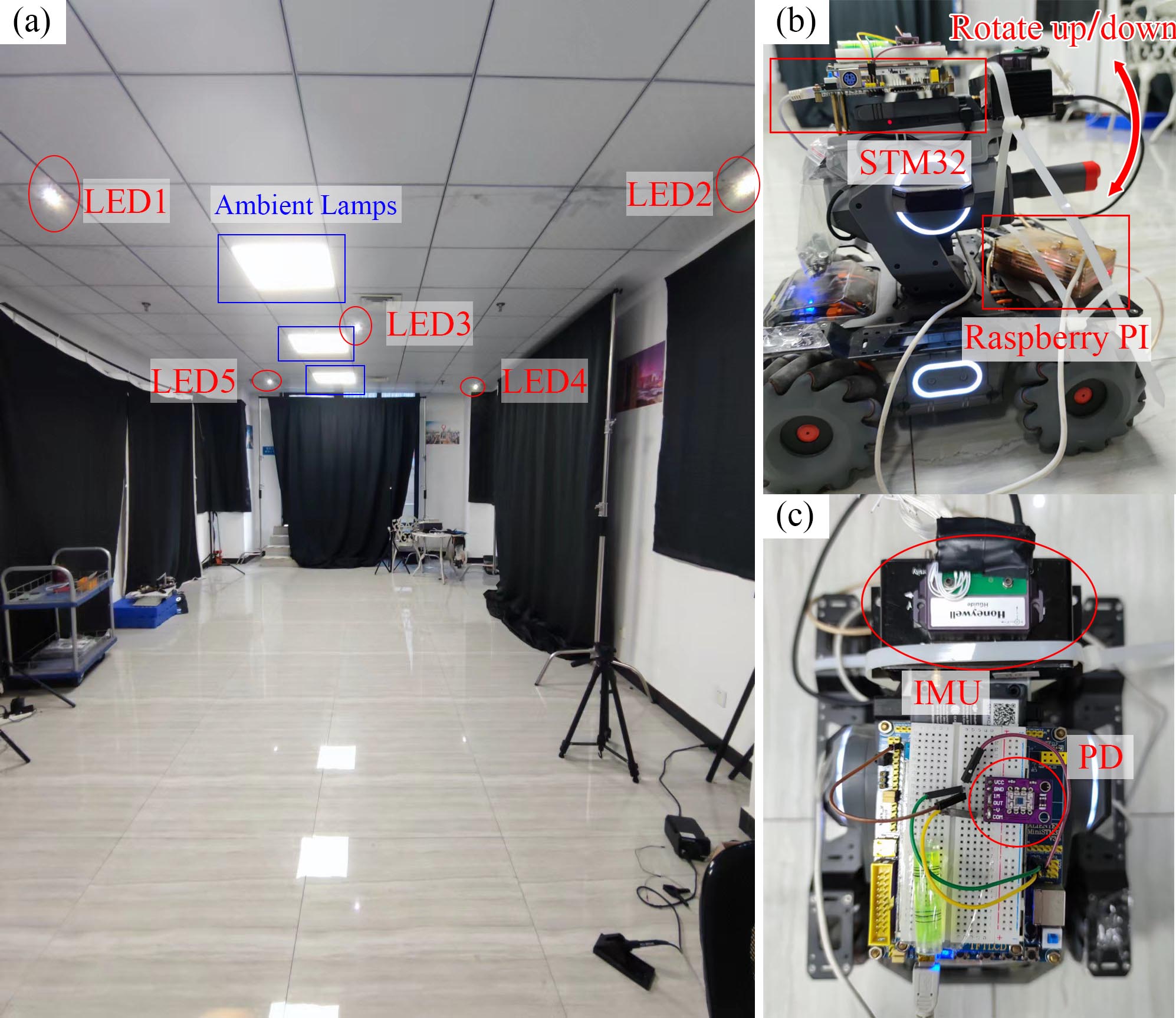}
	\caption{Experimental A's setup. (a) Test site layout; (b) the lateral view of our mobile robot; and (c) the vertical view of our mobile robot. The gimbal of our mobile robot can be rotated up and down.}
	\label{fig_1}
\end{figure}

Experiment A is an ablation study to evaluate the 2D positioning and inclination estimation accuracy by strictly controlling the inclinations and blockages during each test. We implemented several field vehicular tests with five LEDs (CREE XLamp XM-L) mounted on the ceiling as the light beacons with coordinates of (0.35, 1.34), (3.56, 1.15), (1.71, 3.31), (3.50, 6.25), (0.35, 5.97), respectively (Unit: m, shown in Fig. \ref{fig_1}(a)). Each lamp was modulated at a different frequency and controlled by the STM32 Microcontroller Units (MCU). The LEDs were modulated at 1.8 kHz, 2.5 kHz, 3.2 kHz, 3.75 kHz, and 5 kHz, respectively, and the whole transmit power was 10 W. To prevent multipath effects that can severely damage the RSS measurements, we used several pieces of black cloth to block the white walls.

\begin{figure}[H]
	\centering
	\includegraphics[width=3in]{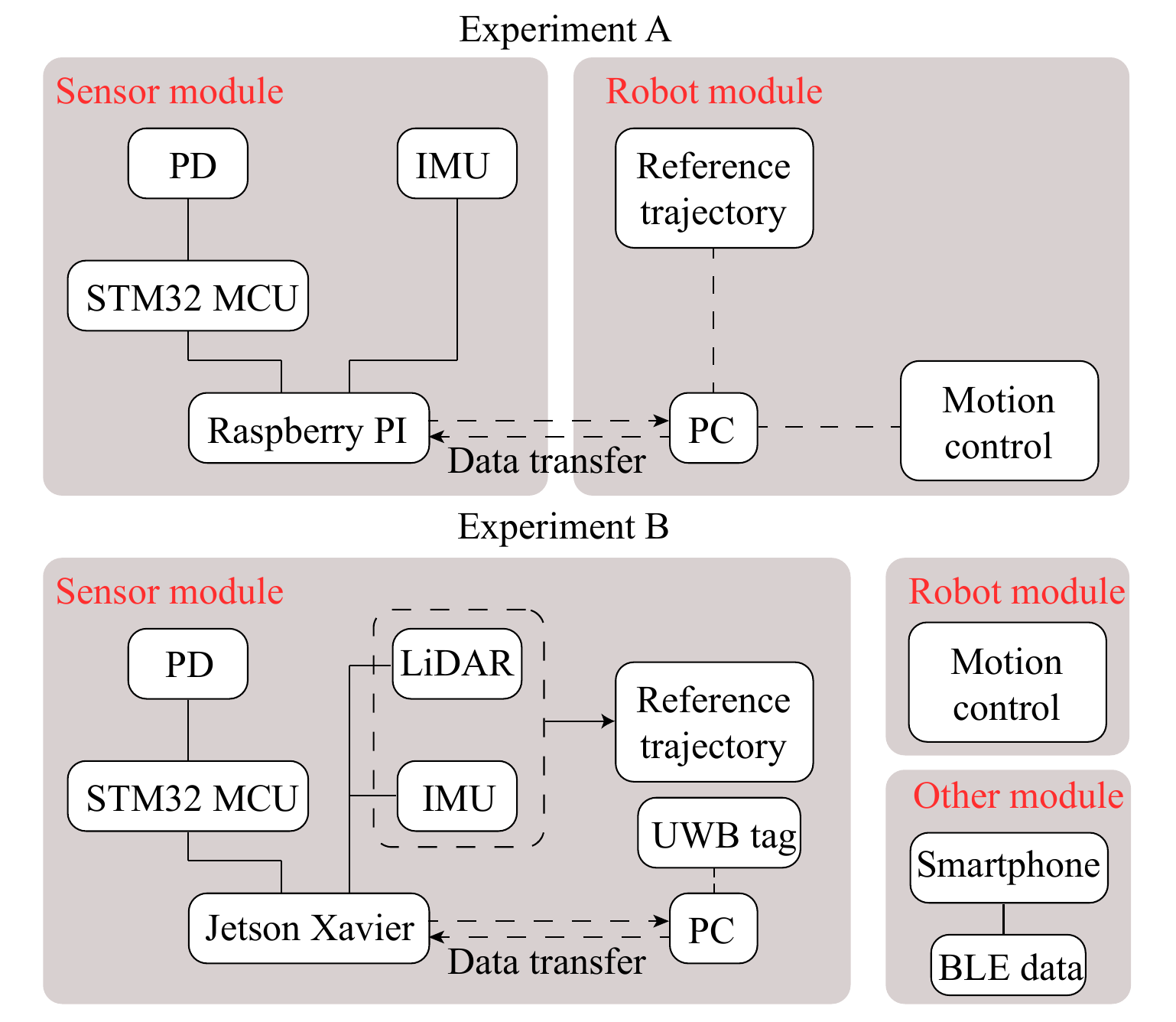}
	\caption{Block diagram of the sensor and the robot system. The dotted lines are wireless connections while the solid lines are wired connections.}
	\label{fig_2}
\end{figure}

The vehicle we used was a mobile robot typed DJI RoboMaster EP. The light sensor, a single OPT101 PD, together with an IMU, was mounted on the rotatable gimbal of the indoor mobile robot, shown in Fig. \ref{fig_1}(b) and \ref{fig_1}(c). The IMU is typed Honeywell HGUIDE i300, whose performance parameters are listed in Table \ref{table1}. Fig. \ref{fig_2} shows the hardware connections, where an STM32 MCU is programmed to make the PD work. The MCU and an IMU were controlled by a Raspberry PI 4B, a credit-card-sized computer whose OS is based on Linux. Our IMU and VLP systems are synchronized based on timestamps given by the Raspberry PI. To vary the inclination of the PD, the gimbal can be rotated to change its pitch angle (shown in Fig. \ref{fig_1}(b)). To test the performance on a fixed track, we called the RoboMaster Software Development Kit (SDK) Application Program Interface (API) using a PC to control motors based on a fixed program and to subscribe to the robot position, which served as a reference for further accuracy evaluation.

To verify the suitability of using the subscribed position for the ground truth, we did 3 groups of rectangular and 2 groups of circular loop closure tests to evaluate the accuracy of robot positions. The lengths of these trajectories are close to the lengths in experiment A. In each test, we measured the displacement between the start and end points of the trajectory; then we compared this displacement to that given by our subscribed positions. Rectangular tests show errors of $3.2\mathrm{cm}$, $3.9\mathrm{cm}$, and $2.3\mathrm{cm}$ and circular tests show errors of $1.2\mathrm{cm}$ and $1.3\mathrm{cm}$. Since the RoboMaster EP itself was equipped with an odometer and a camera, which are both relative positioning sources \cite{ZHUANG202362}, the position errors must be accumulating during motion. Thus, the averaging positioning error must be lower than the loop closure error, which is $2.4 \mathrm{cm}$ by our tests.

\begin{table}
	\caption{Technical characteristics of Honeywell HGUIDE i300 and Xsens MTI-3 from datasheet.}\label{table1}
	\centering
	\begin{tabular}{ccc}
		\hline
		\rule{0pt}{9pt}
		\textbf{Characteristics} & \textbf{HGUIDE i300} & \textbf{MTI-3}\\
		\hline
		Sampling rate & $200 \mathrm{Hz}$ & $100 \mathrm{Hz}$\\
		
		Velocity Random Walk & $0.03 \mathrm{m/s/\sqrt{hr}}$ & $0.04 \mathrm{m/s/\sqrt{hr}}$\\
		
		Angle Random Walk & $0.25^{\circ}/\mathrm{\sqrt{hr}}$ & $0.18^{\circ}/\mathrm{\sqrt{hr}}$\\
		
		Bias Instability of Acce. & $0.03\mathrm{mg}$ & $0.04\mathrm{mg}$\\
		
		Bias Instability of Gyro & $5^{\circ}/\mathrm{hr}$ & $6^{\circ}/\mathrm{hr}$\\
		\hline
	\end{tabular}
\end{table}

Experiment B is to evaluate the performance in a large space and to compare VLP with other indoor positioning systems (UWB and Bluetooth). To our knowledge, this is the largest VLP indoor test site using the range-based positioning algorithms in the VLP literature. The receiver and transmitters of the VLP system were the same in experiment A while the mobile robot and the IMU are different. To effectively cover the whole site, we installed 6 LED lamps of the same type as in experiment A. LED lamps are modulated at 1.8 kHz, 2.5 kHz, 3.2 kHz, 3.75 kHz, 4.35 kHz, and 4.75 kHz, and the modulation methods are the same as experiment A. Seven Bluetooth beacons (SKYLAB VG05) and four UWB tags (LinkTrack P-B) are also installed on this site to transmit wireless signals for indoor positioning. 

\begin{figure}[H]
	\centering
	\includegraphics[width=3.5in]{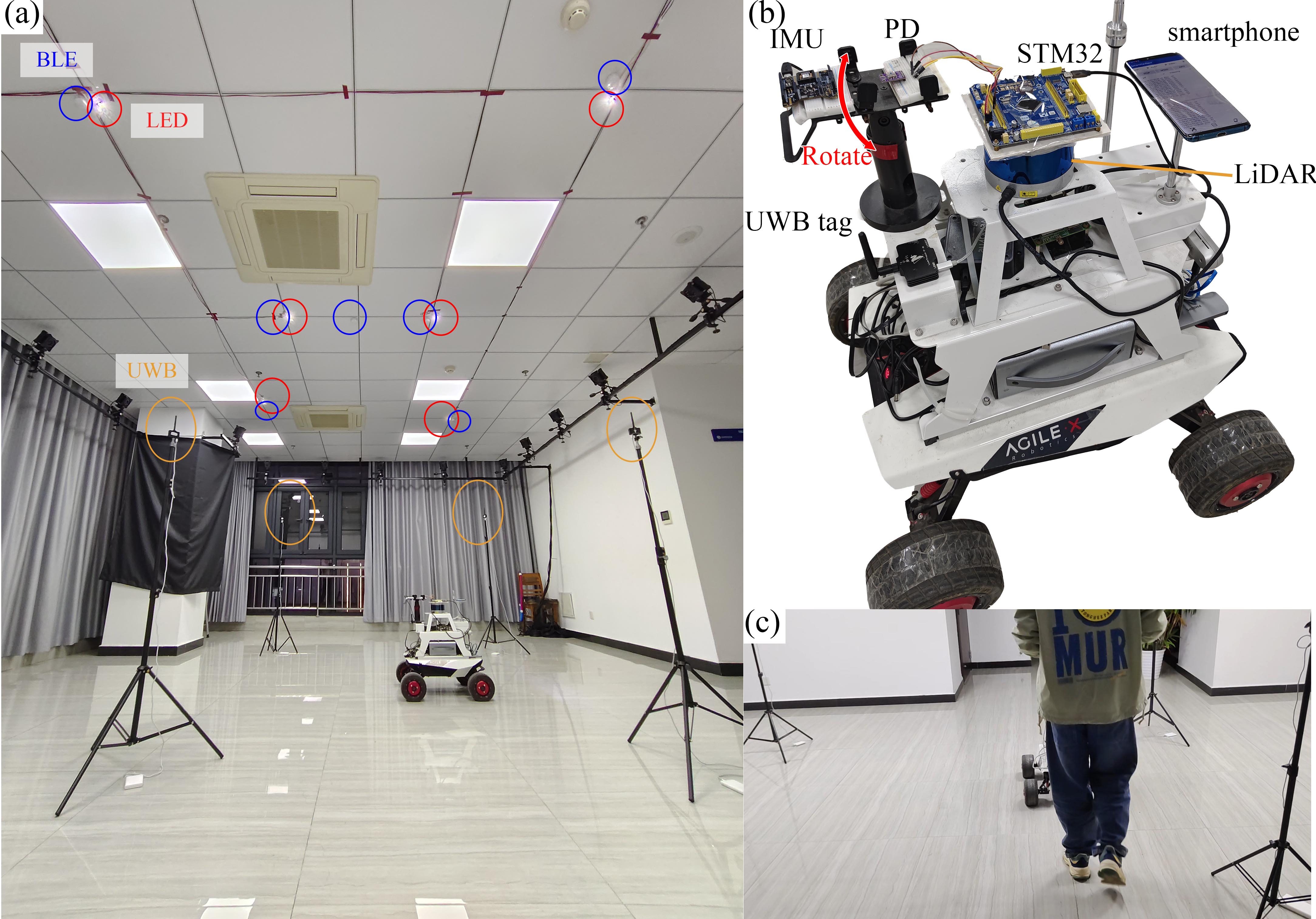}
	\caption{Experimental B's setup. (a) Test site layout: LEDs are circled in red, UWB tags are circled in orange, and BLE beacons are circled in blue; (b) the aerial view of our mobile robot; and (c) the robot operator was blocking the wireless and light signals.}
	\label{exp2}
\end{figure}

The mobile robot we used was an AgileX scout mini, on which a LiDAR (Velodyne VLP16), an IMU (Xsens MTI-3), a UWB tag (LinkTrack P-B), and a smartphone (HUAWEI Mate 20 Pro) are equipped (shown in Fig. \ref{exp2}): the LiDAR provide ground truth positions by integrated with the IMU using the FAST-LIO2 \cite{9697912} program; the smartphone is used to collect Bluetooth Low Energy (BLE) data, together with UWB data to compare with VLP. To test the performance of solving inclinations, we installed the PD and the IMU on a rotatable bracket (shown in Fig. \ref{exp2}). Fig. \ref{fig_2} shows the hardware connections, where we used a Jetson Xavier Developer Kit to control the sensors and collect VLP, IMU, LiDAR, and UWB data. The IMU characteristics are presented in Table \ref{table1}.

In our study, the Ceres solver \cite{ceres-solver} is used to solve the graph optimization problem.

\section{Results and Analysis}
\label{result}
We implemented a simulation test (Section \ref{simulation}), experiment A (Sections \ref{result1}, \ref{result2}, and \ref{result3}), and experiment B (Sections \ref{large}, \ref{unknown}, and \ref{comparison}). In experiment A, we ran two fixed trajectories (rounded rectangular and S-shape trajectory) three times each trajectory: (1) normal tests without inclination and blockage; (2) tests with inclination, and (3) tests with blockages. In experiment B, we test the inclination estimation and blockage detection on a large space, estimate unknown LEDs, and compare VLP with other indoor positioning sensors.

\subsection{Simulation}
\label{simulation}

To test the correctness of the 3D pose estimation and blockage detection algorithm, we built a simulation environment with sloped floors to create height changes. The simulated robot moved uphill for a while and the PD suffered several times blockages. To make the data as realistic as possible, we added while and flicker noises to IMU data whose level is 5 times higher to Table \ref{table1} and white noise to VLP RSS data (1-$\sigma$ strength of 0.1 lux). The size of the whole scenario is $5 \mathrm{m} \times 5 \mathrm{m} \times 5 \mathrm{m}$ and the start point is at (5, 0, 0) (m).

\begin{figure}
	\centering
	\includegraphics[width=3.2in]{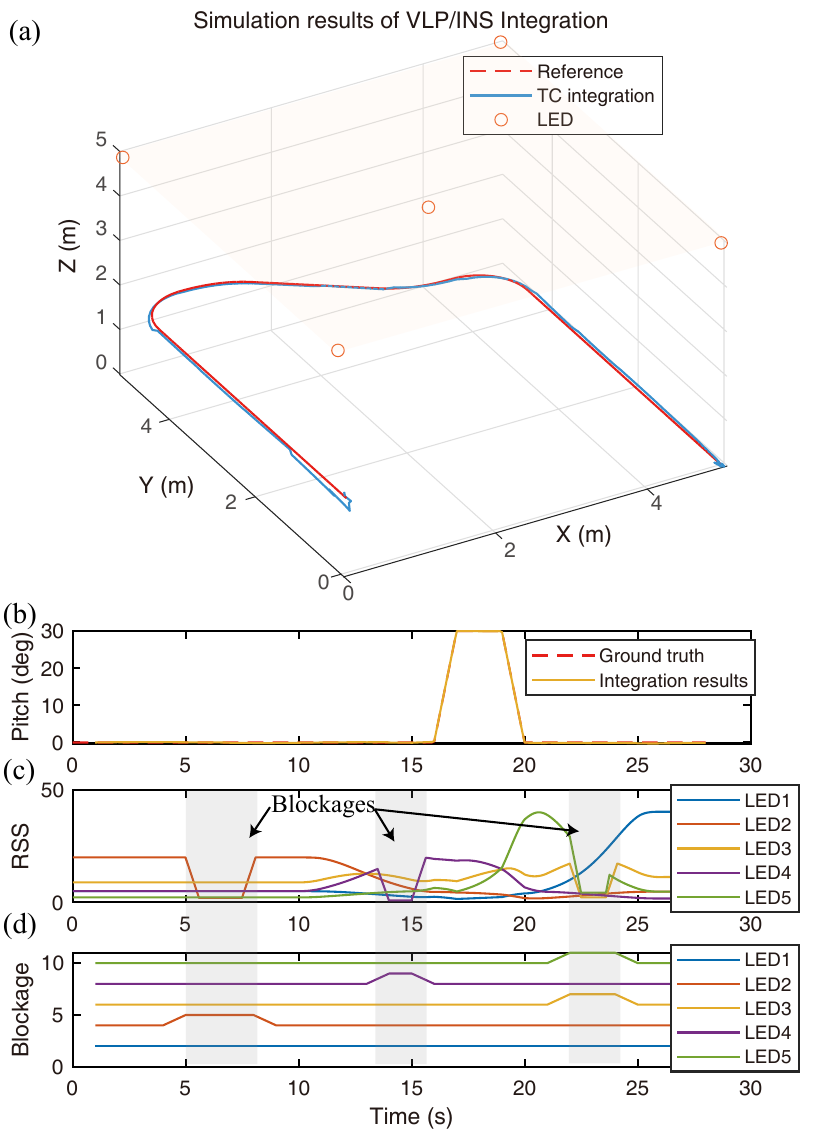}
	\caption{3D simulation results processed by our graph optimization system. (a) Tightly-coupled integration results compared to the referenced trajectory, locations of LEDs are also plotted; (b) pitch angle solved by our system and its ground truth; (c) RSS measurements suffering from blockages; and (d) blockages detected by our DRD method.}
	\label{simu}
\end{figure}

Results in Fig. \ref{simu} show that our system behaves well when estimating the inclination angle and detecting the blocked RSS measurements. The grayed boxes in Fig. \ref{simu}(c) and \ref{simu}(d) show the periods when blockages happened. Fig. \ref{simu}(d) shows the blockages judged by our DRD method; in each curve, an odd number in the y-axis means blockages while even numbers are LOS signals; these values are down-sampled to 1 Hz during the positioning process since RSS measurements are 1 Hz. The average 3D positioning error is $6.2 \mathrm{cm}$ and the average inclination error is $0.08^{\circ}$.

\begin{figure*}
	\centering
	\includegraphics[width=5.2in]{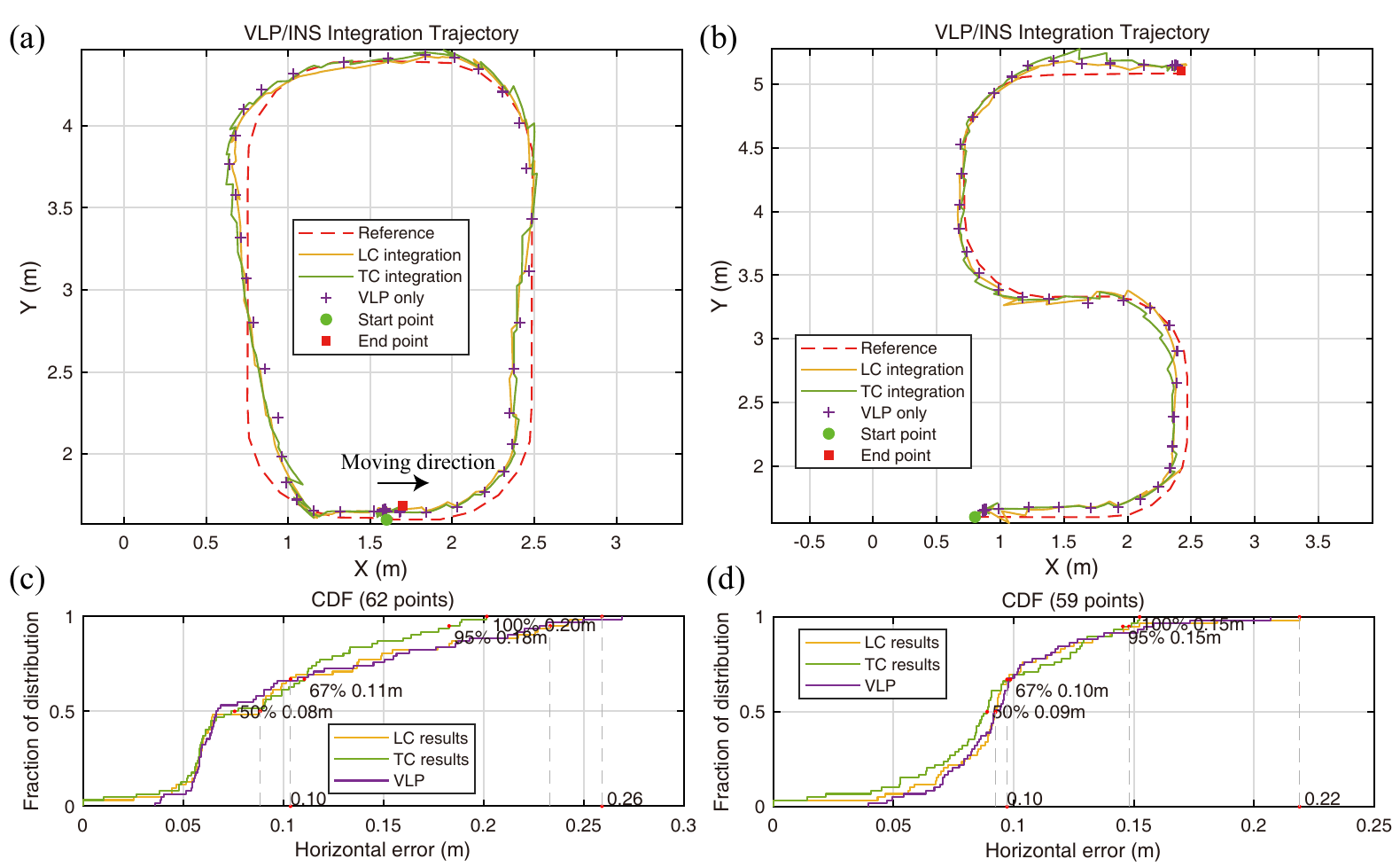}
	\caption{Normal tests' results comparison among Tightly-Coupled (TC) integration, Loosely-Coupled (LC) integration, and VLP alone. (a) Rounded rectangular trajectory; (b) S-shape trajectory; and (c)-(d), their CDFs.}
	\label{normal_plot}
\end{figure*}
\subsection{Normal Tests without Inclination and Blockage}
\label{result1}
Normal tests are control groups to compare with the inclination tests (Section \ref{result2}) and blockage tests (Section \ref{result3}); in these tests, we compare our Tightly-Coupled (TC) system with Loosely-Coupled (LC) VLP/INS and VLP alone. As Fig. \ref{tight_loose} shows, the LC system uses the integration of IMU data to determine the attitude and solves the positions using RSS data. Different from our TC system where raw RSS measurements are fused with IMU data, the LC system fuses the positions with IMU data. As another control group, the VLP-alone system does not use IMU data: in the test without inclinations, we assumed inclinations as zero and simplified the Lambertian model (\ref{lamber2}) as:
\begin{equation}
	P_l=\frac{\left(m_l+1\right) A_R P_{Tl}}{2 \pi} \cdot \frac{h^2_l} {D^{3+m_l}_l}
\end{equation}
where $h_l$ is the constant height difference between the $l^{th}$ LED and the PD. Distance ${D_l}$ can be directly derived from RSS measurement $P_l$.

We used the positions given by DJI RoboMaster's SDK API as the reference (red dotted curves in Fig. \ref{normal_plot}), whose sampling rate is 1Hz. The startpoint, endpoint, and the initial moving direction are plotted on the graph. We also calculated the cumulative distribution functions (CDFs, Fig. \ref{normal_plot}(c) and \ref{normal_plot}(d)) based on that reference. To evaluate the positioning errors, we calculated the distances between solved positions and referenced positions. Trajectories and CDFs show that the accuracy of the TC, LC, and VLP-alone results is similar. 

\subsection{Tests with Inclination}
\label{result2}
\begin{figure*}
	\centering
	\includegraphics[width=5.5in]{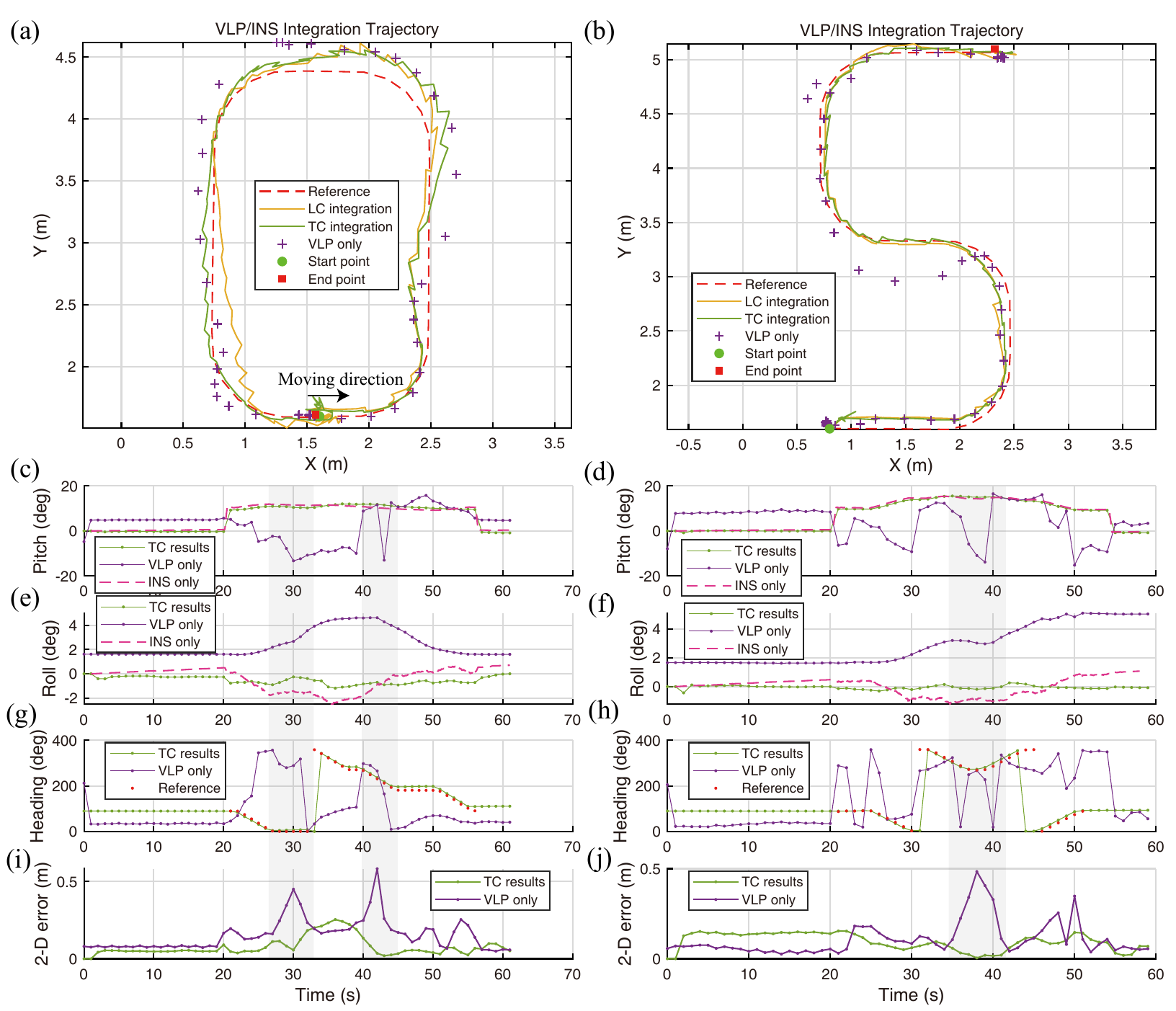}
	\caption{Inclination tests' results comparison among TC, LC integration, and VLP alone. The gray boxes indicate the periods when both pitch angle errors and position errors are large. (a) Rounded rectangular trajectory; (b) S-shape trajectory; (c) pitch angles in rounded rectangular trajectory; (d) pitch angles in S-shaped trajectory; (e) heading angles in rounded rectangular trajectory; (f) heading angles in S-shaped trajectory; (g) roll angles in rounded rectangular trajectory; (h) roll angles in S-shaped trajectory; (i) 2-D errors in rounded rectangular trajectory; and (j) 2-D errors in S-shaped trajectory.}
	\label{inclination_plot}
\end{figure*}

Our system optimizes positions and inclinations simultaneously within a graph optimization frame. To test our system's ability to deal with inclination changes, we ran the same trajectories as Section \ref{result1} does but changed the gimbal's inclination. In these inclination tests, we called an API function of our robot to rotate its gimbal to a pitch angle of 10 deg during the movement periods. During the motion of our robot, rapid turnings made the inclination change; thus, pitch and roll angles were not fixed. However, the roll angles were small and could be assumed as zero. Due to the introduction of the inclination, the VLP-alone system should additionally solve pitch and heading angles, where we used the propagation model (\ref{lamber2}) and assumed: 
\begin{equation}
	\begin{aligned}
		\mathbf{n}^u&= \left[\begin{matrix}\mathrm{cos}(\theta_z)\mathrm{sin}(\theta_y) & \mathrm{sin}(\theta_z)\mathrm{sin}(\theta_y) & \mathrm{cos}(\theta_y)\end{matrix}\right]^T\\
		\mathbf{D}_l^u&=\left[\begin{matrix}x_l & y_l & h_l\end{matrix}\right]^T
	\end{aligned}
\end{equation}
where $\theta_z$ is the heading, $\theta_y$ is the pitch angle of the gimbal we set, $x_l$ and $y_l$ compose the 2-D position difference. $\theta_z$, $\theta_y$, $x$, and $y$ are variables to be optimized while $h_l$ is the constant height difference. During movement, the true values of roll and pitch angles were hard to solve; but in the static phase, they were accurately measured: roll angles are zero, and pitch angle differences are shown in Table \ref{table2}. The true values of heading angles during movement were calculated based on the 2-D ground truth positions (red dot in Fig. \ref{inclination_plot}(g) and \ref{inclination_plot}(h)).

The TC, LC, and VLP-alone results are shown in Fig. \ref{inclination_plot}. The accuracy of TC results in Fig. \ref{inclination_plot}(a) in the lower left corner of the trajectory surpasses LC results due to the attitude errors accumulated from the gyroscope integration. In both Fig. \ref{inclination_plot}(a) and \ref{inclination_plot}(b), VLP-alone results deviate much from the ground truth since the VLP-alone system can hardly solve the attitudes as the IMU-aided TC and LC systems do. This conclusion can be drawn in Fig. \ref{inclination_plot}(c)-\ref{inclination_plot}(h): compared with the attitudes (pitches, rolls, and headings) solved by the TC system (green curves), those variables solved by the VLP-alone system (purple curves) are unreliable. Especially, heading angles (purple curves in Fig. \ref{inclination_plot}(g) and \ref{inclination_plot}(h)) are completely wrong, which is consistent with our conclusions drawn in Section \ref{disturbance} that heading angles are not observable. In Fig. \ref{inclination_plot}(i) and \ref{inclination_plot}(j), we plot the 2D-error series calculated based on the referenced trajectory. We use gray boxes to mark the periods when the VLP-alone system performs worst. In these boxes, pitch angle errors are extremely large. Overall, the wrongly solved inclination angles lead to large positioning errors in the VLP-alone system.

To further explore the inclinations' accuracy, we compare them to the results of INS mechanization \cite{RN20}. Fig. \ref{inclination_plot}(c)-\ref{inclination_plot}(f) shows the inclinations solved by our tightly-coupled VLP/INS system (blue curves) and the INS mechanization results (red curves). Since the RoboMaster API cannot provide accurate gimbal attitudes as the referenced position does, we manually measured the two pitch angles before and after each kinematic test. The differences between them are used as a reference to verify the attitudes solved by our VLP/INS fusion system. Table \ref{table2} shows the attitudes of VLP/INS integration are more accurate than the pure INS. Due to the IMU bias, scale factors, and some random error, pure INS results gradually deviate from true inclination.

\begin{table}
	\caption{The pitch angle difference during each test.}\label{table2}
	\centering
	\begin{tabular}{cccc}
		\hline
		\rule{0pt}{9pt}
		\textbf{Tests} & \textbf{Measured} & \textbf{VLP/INS} & \textbf{Pure INS} \\
		\hline
		Rounded rectangular trajectory & $-0.40^{\circ}$ & $-0.50^{\circ}$ & $0.39^{\circ}$\\
		S-shaped trajectory & $-0.83^{\circ}$ & $-0.85^{\circ}$ & $-0.39^{\circ}$\\
		\hline
	\end{tabular}
\end{table}

\subsection{Tests with Blockages}
\label{result3}
\begin{figure*}
	\centering
	\includegraphics[width=5.5in]{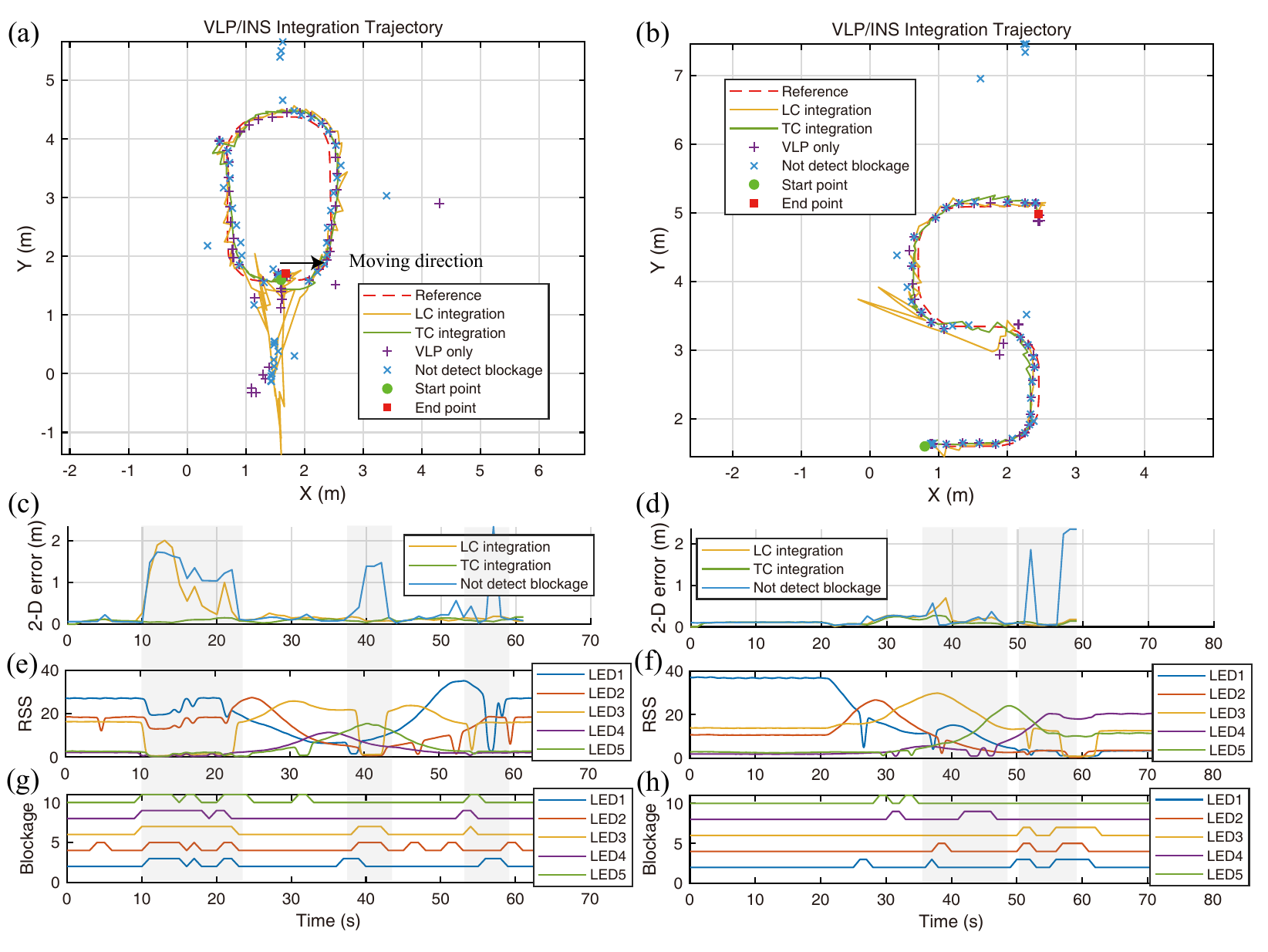}
	\caption{Blockage tests' results comparison among TC integration, LC integration, and VLP alone. The gray boxes indicate the periods when light blockages are severe. (a) Rounded rectangular trajectory; (b) S-shape trajectory; and (c)-(d), their 2-D errors; (e) the blocked RSS in rounded rectangular trajectory; (f) the blocked RSS in S-shaped trajectory; (g) the detection of blockages in rounded rectangular trajectory; and (h) the detection of blockages in S-shaped trajectory.}
	\label{blockage_plot}
\end{figure*}

A key advantage of our VLP/INS system is the ability to deal with light blockages. Blockages lead to the loss of RSS measurements, therefore, VLP alone sometimes can hardly work out reliable solutions while the TC VLP/INS system can still work out smooth solutions with the dead reckoning of INS. To test our system's ability to deal with blockages, we ran the same trajectories as Section \ref{result1}. In these tests, a pedestrian walked randomly in the room to block the visible light signals. Fig. \ref{blockage_plot} shows two tests with blockages, where TC, LC, and VLP-alone results in Fig. \ref{blockage_plot}(a) and \ref{blockage_plot}(b) all adopt the blockage detection strategy (i.e., DRD) as we demonstrated in Section \ref{blockage} to exclude blocked measurements. Due to the severe damage of blockages, positions solved by VLP alone are not successive and contain large errors. Since the LC system adopts VLP-alone positioning results as inputs, large oscillations occur in the LC navigation trajectory (some even larger than 1 meter, shown in Fig. \ref{blockage_plot}(c) and \ref{blockage_plot}(d)).

To validate the need for detecting blockages, we designed a control group using VLP data only (blue points in Fig. \ref{blockage_plot}(a)-\ref{blockage_plot}(d)) where we did not detect blockages to pretend all measurements are effective. Without the blockage detection strategy, positioning accuracy is extremely bad, especially for the parts in gray boxes.

To verify the effectiveness of our blockage detection method, we plot the RSS variations and detection results in Fig. \ref{blockage_plot}(e)-\ref{blockage_plot}(h). Fig. \ref{blockage_plot}(e) and \ref{blockage_plot}(f) are the RSSs sampled with a rate of 120 Hz, which present some sudden reductions when blockages happen. Fig. \ref{blockage_plot}(g) and \ref{blockage_plot}(h) are the blockages judged by our DRD method; in each curve, an odd number in the y-axis means blockages while even numbers are LOS signals; these values are down-sampled to 1 Hz during the positioning process since RSS measurements are 1 Hz. Results show that the detection accuracy is 100\%.

In conclusion, Table \ref{table3} presents the accuracy comparison of three groups of tests discussed in the above three subsections. Our tightly-coupled system gets an average accuracy of 9.6 centimeters even with NLOS and PD tilting.
\begin{table}
	\caption{Accuracy comparison among three groups of tests (Unit: meter).}\label{table3}
	\centering
	\setlength{\tabcolsep}{1pt}
	\begin{tabular}{ccccccc}
		\hline
		\multirow{2}{*}{\textbf{Tests}} & \multicolumn{2}{c}{\textbf{TC VLP/INS}} & \multicolumn{2}{c}{\textbf{LC VLP/INS}} & \multicolumn{2}{c}{\textbf{VLP alone}} \\& Rectangular & S-shape & Rectangular & S-shape & Rectangular & S-shape \\
		\hline
		Normal tests & $0.102$ & $0.095$ & $0.116$& $0.100$ & $0.113$ & $0.099$\\
		Inclination tests & $0.088$ & $0.088$ & $0.149$& $0.097$ & $0.178$ & $0.126$\\
		Blockage tests & $0.094$ & $0.111$ & $0.285$& $0.140$ & $0.331$ & $0.146$\\
		\hline
	\end{tabular}
\end{table}

\subsection{Large Scale Tests}
\label{large}
\begin{figure}
	\centering
	\includegraphics[width=2.8in]{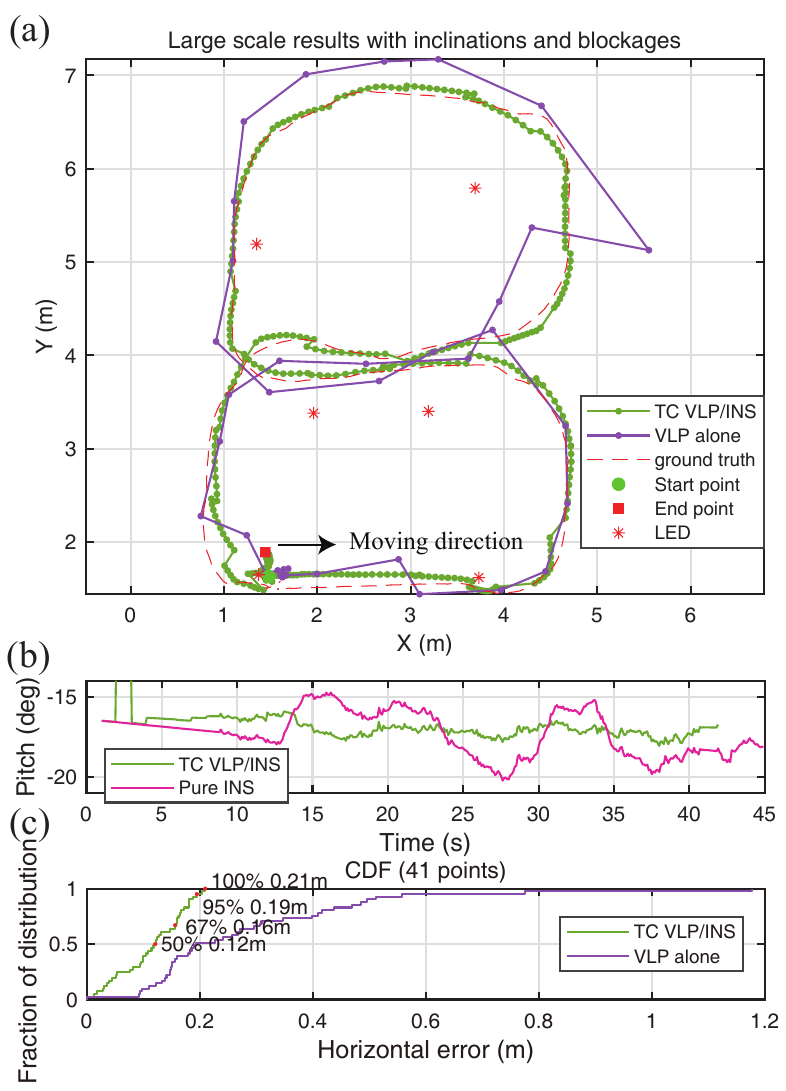}
	\caption{Large space test. (a) The trajectory of VLP/INS integration compared with VLP-alone and ground truth; (b) the pitch angles calculated by TC VLP/INS integration and pure INS; and (c) the CDFs.}
	\label{large_integrate}
\end{figure}

In this section, we test our system in a large space under blockages and inclinations. The trajectories, pitch angles, and CDFs are given in Fig. \ref{large_integrate}. In Fig. \ref{large_integrate}(b), we compare the pitch angles calculated by graph optimization and pure INS; the pitch angles calculated by VLP alone are not plotted since they are inaccurate (analyzed in Section \ref{result1}). The initial and final pitch angles were measured accurately at $16.5^{\circ}$; but during movement, the pitch angle may fluctuate up and down this value. At the end of this trajectory, the pitch angle of TC VLP/INS deviated by $0.2^{\circ}$ but that angle calculated by the integration of the gyroscope deviated by $1.6^{\circ}$ (Fig. \ref{large_integrate}(b)). The CDFs in Fig. \ref{large_integrate}(c) compare the 2D positioning results of TC VLP/INS integration and pure VLP based on the ground truth calculated using LiDAR and IMU through the FAST-LIO2 program. Since VLP alone can hardly solve inclinations and headings (discussed in Section \ref{result1}), its accuracy is far worse than that of the integration results. We achieved a mean accuracy of 0.115 m and the largest error was 0.209 m, which is similar to experiment A. This result verifies the robustness in a large space on the situation of PD inclination and signal blockages. 

\subsection{Estimating Unknown LEDs}
\label{unknown}
In this section, we set some LEDs' locations to be unknown and use Section \ref{large}'s data to simultaneously estimate the trajectory and the unknown LED locations. Each LED's initial location was randomly set and the sliding window size was set to be a large value (50 in our test). From Table \ref{unknownLED}, when only one LED's location was unknown, the locations of LEDs 3, 4, 5, and 6 were accurately estimated during the optimization, while the locations of LEDs 1 and 2 were not. This phenomenon can be interpreted using our hypothesis in Section \ref{sec_unknown}: based on Fig. \ref{large_integrate}, for example, LED 1's DOP value is high since it is close to the edge of the trajectory, while LED 4's DOP is relatively low. Tests with two unknown LEDs also validate this idea. But when three LEDs' locations were unknown, the missing of some critical LEDs (3, 4, 5, or 6) may lead to the divergence of our optimizer. In the last group of Table \ref{unknownLED}, none of the three LEDs' locations converged to stable values. However, several LEDs' missing locations did not lead to a significant decrease in the trajectory's accuracy. In general, the accuracy of estimated LED locations and that of the trajectory are positively correlated.
\begin{table}
	\caption{Accuracy of the trajectories and LEDs when some LEDs' locations are unknown (Unit: meter).}\label{unknownLED}
	\centering
	\begin{tabular}{m{1.6cm}<{\centering}m{1.6cm}<{\centering}m{1cm}<{\centering}m{1cm}<{\centering}m{1cm}<{\centering}}
		\hline
		\textbf{Unknown LEDs} & \textbf{Trajectory mean error} & \multicolumn{3}{c}{\textbf{LEDs' location errors}}\\
		\hline
		LED 1 & $0.170$ & $0.107$ & -& -\\
		LED 2 & $0.150$ & $0.149$ & -& - \\
		LED 3 & $0.144$ & $0.018$ & -& - \\
		LED 4 & $0.128$ & $0.022$ & -& - \\
		LED 5 & $0.118$ & $0.034$ & -& - \\
		LED 6 & $0.124$ & $0.064$ & -& - \\
		LEDs 2,4 & $0.166$ & $0.153$ & $0.028$ & - \\
		LEDs 3,6 & $0.157$ & $0.016$ & $0.065$ & - \\
		LEDs 2,4,6 & $0.177$ & $0.154$ & diverge & $0.070$ \\
		LEDs 1,2,5 & $0.128$ & $0.077$ & $0.106$ & $0.046$ \\
		LEDs 3,4,6 & $0.197$ & diverge & diverge & diverge \\
		\hline
	\end{tabular}
\end{table}

\subsection{Comparing VLP with Other IPSs}
\label{comparison}
\begin{figure}
	\centering
	\includegraphics[width=2.8in]{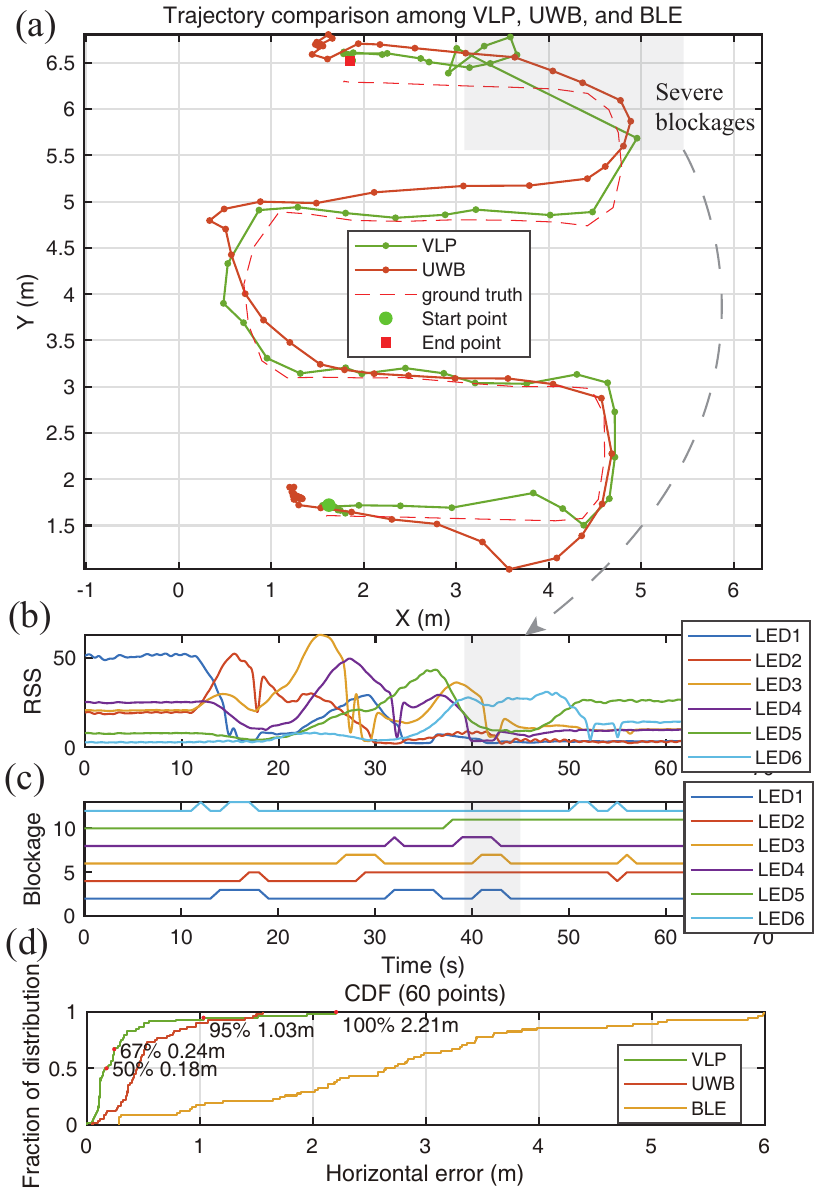}
	\caption{Comparing VLP with UWB and BLE. (a) The VLP and UWB's trajectories compared with ground truth positions; (b) the blocked RSS measurements; (c) the judgment of blockages; and (d) the CDFs of VLP, UWB, and BLE.}
	\label{VLP_UWB_plot}
\end{figure}

We plot positioning results of pure VLP, UWB, and BLE to compare with each other. In this test, the robot operator blocked LOS signals and severely disturbed wireless and light signals as Fig. \ref{exp2}(c) shows. PD was horizontally placed since the pure VLP can hardly solve the inclinations and the heading angle. Data were preprocessed in two steps: all positioning results were down-sampled to 1 Hz and synchronized to the same periods; lever-arm corrections were used to make them aligned to the same center point. We used our blockage detection method to effectively suppress NLOS effects in VLP (results are shown in Fig. \ref{VLP_UWB_plot}(b) and \ref{VLP_UWB_plot}(c)), while NLOS in UWB and BLE were not suppressed since they are radio-based signals and the NLOS problem among them is hard to solve. Fig. \ref{VLP_UWB_plot}(a) shows their trajectories' comparison, and Fig. \ref{VLP_UWB_plot}(d) shows their CDFs. BLE does not appear in Fig. \ref{VLP_UWB_plot}(a) because its trajectory is too disorganized. These figures show that VLP is much more accurate unless light signals are severely blocked, which causes RSS measurements insufficient to locate (a large error occurs in the upper-right corner of Fig. \ref{VLP_UWB_plot}(a)). The accuracy differences are reasonable because, during LOS situations, the ranging accuracy of VLP (within 10 cm) is higher than UWB (worse than 10 cm) in our experiment and much higher than BLE (worse than 2 m). 

\section{Conclusion}
\label{conclusion}
In this paper, we proposed a tightly coupled VLP/INS integrated navigation system to solve the problem of dynamic inclination estimation and blockage elimination. Innovatively, our system analyzes the observability of the attitude using the disturbance model of Lambertian law and detects the blocked observations based on the RSS changing rate. We also prove the feasibility of simultaneously estimating PD's pose and some unknown LEDs' locations. To verify the effectiveness, we did simulations and implemented two groups of real-world experiments using mobile robots assembled with low-cost PD and IMUs. Under the condition of dynamic inclination changes and light blockages, our experiments achieve an overall accuracy of around 10 cm during movement and inclination accuracy within 1 degree. Due to the widespread use of LEDs and the miniaturization of PDs and their PCBs for control, our proposed system is suitable for consumer and industrial applications, especially for mobile robots, mechanical arms, wearable devices, etc.

\section*{Declarations}

\bmhead{Acknowledgments}
The authors would like to acknowledge Prof. Xiaoji Niu and the Integrated and Intelligent Navigation (i2Nav) group from Wuhan University for providing the OB\_GINS software and Honeywell HGUIDE i300 IMU that were used in this research. We also acknowledge Tengfei Yu, Xuan Wang, and Xiaoxiang Cao for their contribution to the experiments.

\bmhead{Data availability}
The datasets used and/or analysed during the current study are available from the corresponding author on reasonable request.

\bmhead{Competing interests}
The authors declare that they have no competing interests.

\bmhead{Authors' contributions}
Y.Z proposed the idea of tightly integrating VLP and INS and revised the manuscript. X.S is responsible for formula derivation, experimental calculation, analysis and writing the draft manuscript. X.Y and T.H participated in the experiments. J.H provided program guidance and revised the manuscript. All authors read and approved the final manuscript.

\bmhead{Funding}
This work was supported by Excellent Youth Foundation of Hubei Scientific Committee, China (2021CFA040).

\begin{appendices}
\section{Disturbance of Lambertian model}\label{secA1}
Based on the error disturbance \cite{RN20} of DCM ${\mathbf{R}}_v^u$ with respect of a small angle $d\boldsymbol{\phi}=d\boldsymbol{\phi}^u_{vu}$,
\begin{equation}
	\begin{aligned}
		\hat{\mathbf{R}}_v^u&={\mathbf{R}}_{v^{\prime}}^{u^{\prime}}={\mathbf{R}}_{u}^{u^{\prime}}{\mathbf{R}}_{v}^{u}{\mathbf{R}}_{v^{\prime}}^{v}\\
		&=\left[\mathbf{I}-\left(\boldsymbol{\theta}_{uu^{\prime}}^u \times \right)\right]\mathbf{R}_v^u\left[\mathbf{I}+\left(\boldsymbol{\theta}_{vv^{\prime}}^v \times \right)\right]\\
		&\approx\mathbf{R}_v^u-\left(\boldsymbol{\theta}_{uu^{\prime}}^u \times \right)\mathbf{R}_v^u+\mathbf{R}_v^u\left(\boldsymbol{\theta}_{vv^{\prime}}^v \times \right)\\
		&=\mathbf{R}_v^u-\left(\boldsymbol{\theta}_{uu^{\prime}}^u \times \right)\mathbf{R}_v^u+\left(\boldsymbol{\theta}_{vv^{\prime}}^u \times \right)\mathbf{R}_v^u\\
		&=\left[\mathbf{I}-\left(d\boldsymbol{\phi}^u_{vu} \times \right)\right]\mathbf{R}_v^u=\left[\mathbf{I}-\left(d\boldsymbol{\phi} \times \right)\right]\mathbf{R}_v^u
	\end{aligned}
\end{equation}
we have
\begin{equation}
	\begin{aligned}
		d\mathbf{n}^u &= d\left(\hat{\mathbf{R}}^u_{v_k}\mathbf{n}^{v_k}\right)\\
		&=\left[\mathbf{I}-\left(d\boldsymbol{\phi} \times \right)\right]\mathbf{R}_{v_k}^u\mathbf{n}^{v_k}-\mathbf{R}_{v_k}^u\mathbf{n}^{v_k}\\
		&=-d\boldsymbol{\phi}\times\mathbf{n}^u=\mathbf{n}^u\times d\boldsymbol{\phi}
	\end{aligned}
\end{equation}

Based on equation (\ref{lamber2}), we derive the complete differential considering the position and attitude variables:
\begin{equation}
	\begin{aligned}
		dP_l=&\frac{\left(m_l+1\right) A_R P_{Tl}}{2 \pi} \cdot \\
		&\left[\frac{ \left((\mathbf{n}^u\times d\boldsymbol{\phi})\cdot\mathbf{D}_l^u - \mathbf{n}^u\cdot d\mathbf{r}\right) \left(\mathbf{n}_{l}^u\cdot\mathbf{D}_l^u\right)^{m_l}} {D^{3+m_l}_l}\right.\\
		&-\frac{ m_l\left(\mathbf{n}^u\cdot\mathbf{D}_l^u\right) \left(\mathbf{n}_{l}^u\cdot\mathbf{D}_l^u\right)^{m_l-1} \left(\mathbf{n}_{l}^u\cdot d\mathbf{r}\right)} {D^{3+m_l}_l}\\
		&+\left. \frac{3+m_l}{2} \frac{ \left(\mathbf{n}^u\cdot\mathbf{D}_l^u\right) \left(\mathbf{n}_{l}^u\cdot\mathbf{D}_l^u\right)^{m_l}} {\left(D^{2}_l\right)^{\frac{3+m_l}{2}+1}}\left(2\mathbf{D}_l^u \cdot d\mathbf{r}\right)\right]\\
		=&P_l\frac{(\mathbf{D}_l^u\times\mathbf{n}^u)\cdot d\boldsymbol{\phi}} {\mathbf{D}_l^u\cdot\mathbf{n}^u} + P_l\left[-\frac{\mathbf{n}^u\cdot d\mathbf{r}}{\mathbf{n}^u\cdot\mathbf{D}_l^u} -\frac{m_l\left(\mathbf{n}_{l}^u\cdot d\mathbf{r}\right)}{\mathbf{n}_{l}^u\cdot\mathbf{D}_l^u}\right.\\
		&+\left.\frac{3+m_l}{2}\frac{2\mathbf{D}_l^u \cdot d\mathbf{r}}{D^{2}_l}\right]
	\end{aligned}
\end{equation}

\end{appendices}

\bibliography{reference}
\end{document}